\documentclass[letterpaper]{article} % DO NOT CHANGE THIS
\usepackage[]{aaai25}  % DO NOT CHANGE THIS
\usepackage{times}  % DO NOT CHANGE THIS
\usepackage{helvet}  % DO NOT CHANGE THIS
\usepackage{courier}  % DO NOT CHANGE THIS
\usepackage[hyphens]{url}  % DO NOT CHANGE THIS
\usepackage{graphicx} % DO NOT CHANGE THIS
\def\UrlFont{\rm}  % DO NOT CHANGE THIS
\usepackage{natbib}  % DO NOT CHANGE THIS AND DO NOT ADD ANY OPTIONS TO IT
\usepackage{caption} % DO NOT CHANGE THIS AND DO NOT ADD ANY OPTIONS TO IT
\usepackage{algorithm}
\usepackage{graphicx}
\usepackage{amsmath}
\usepackage{multirow}
\usepackage{algorithm}
\usepackage{algpseudocode}
\usepackage{comment}
\usepackage{xcolor}
\usepackage{amsfonts}

\usepackage{newfloat}
\usepackage{listings}
\DeclareCaptionStyle{ruled}{labelfont=normalfont,labelsep=colon,strut=off} % DO NOT CHANGE THIS
\floatstyle{ruled}
\newfloat{listing}{tb}{lst}{}
\floatname{listing}{Listing}
\title{Towards Transparent Stance Detection: A Zero-Shot Approach Using Implicit and Explicit Interpretability}
\author {
    Apoorva Upadhyaya,
    Wolfgang Nejdl,
    Marco Fisichella
}
\affiliations {
    L3S Research Center, Leibniz Universität Hannover, Hannover, Germany \\
    \{upadhyaya, nejdl, mfisichella\}@l3s.de
}

\usepackage{bibentry}
\begin{document}

\maketitle

\begin{abstract}
Zero-Shot Stance Detection (ZSSD) identifies the attitude of the post toward unseen targets. Existing research using contrastive, meta-learning, or data augmentation suffers from generalizability issues or lack of coherence between text and target. Recent works leveraging large language models (LLMs) for ZSSD focus either on improving unseen target-specific knowledge or generating explanations for stance analysis. However, most of these works are limited by their over-reliance on explicit reasoning, provide coarse explanations that lack nuance, and do not explicitly model the reasoning process, making it difficult to interpret the model’s predictions. To address these issues, in our study, we develop a novel interpretable ZSSD framework, IRIS. We provide an interpretable understanding of the attitude of the input towards the target implicitly based on sequences within the text (implicit rationales) and explicitly based on linguistic measures (explicit rationales). IRIS considers stance detection as an information retrieval ranking task, understanding the relevance of implicit rationales for different stances to guide the model towards correct predictions without requiring the ground-truth of rationales, thus providing inherent interpretability. In addition, explicit rationales based on communicative features help decode the emotional and cognitive dimensions of stance, offering an interpretable understanding of the author's attitude towards the given target. Extensive experiments on the benchmark datasets of VAST, EZ-STANCE, P-Stance, and RFD using ${50\%}$, ${30\%}$, and even ${10\%}$ training data prove the generalizability of our model, benefiting from the proposed architecture and interpretable design.
\end{abstract}


\section{Introduction}
%\textcolor{red}{To improve the introduction you could clarify the problem statement more succinctly upfront. For example, explicitly state the gap in existing ZSSD methods and then immediately tie it to the need for interpretability and generalization.}
Zero-shot stance detection (ZSSD) is about recognizing the author's stance towards the unknown targets. Current advances in ZSSD have mainly focused on adversarial, meta-learning, or data augmentation approaches \citep{zhang2024commonsense,wang2024meta}. However, these techniques suffer from generalization problems due to the uneven distribution of targets or the lack of high-quality annotated training data. In addition, large language models (LLMs) have been used for stance detection (SD) \citep{ding2024edda,zhang2024llm}. Most of the existing studies using LLMs aim to improve target-specific knowledge for the unseen targets with the input text \cite{guo2024improving,zhang2024llm}. Some of the more recent work has focused on interpretable models, such as \cite{zhang2024knowledge} developed a knowledge-augmented interpretable network using LLM to provide perspectives towards targets, \cite{saha2024stance} constructs a stance tree to provide explanations for claim-based stance, and \cite{ding2024distantly} uses instruction-based chain-of-thought with LLMs to generate explanations for stance analysis.
 %\citep{taranukhin2024stance} exploits chain-of-thought in context learning by using explicit reasoning using LLM. 
\par While these works provide a high-level perspective, they often lack fine-grained insight into the word-level influences that may affect the overall decision-making process for identifying stance, thus reducing interpretability at a detailed level. Moreover, most of these studies rely on explicit inference or acquisition of target-specific knowledge, which can overlook nuanced or implicit cues in the input text, especially subtle attitudes. To address these issues, in contrast to existing works that are limited by over-reliance on coarse explanations \cite{saha2024stance, ding2024distantly}, we develop an interpretable ZSSD system that incorporates both implicit rationales based on sequences within the text and explicit rationales based on linguistic measures. This provides a more fine-tuned, interpretable method that enables a more comprehensive understanding of the author's stance by capturing both subtle and overt cues, ultimately improving both transparency and granularity.

%One of the works utilizes LLM to explicitly extract the relationship between paired text and target \cite{zhang2024llm}. Some of the works focused on explanaing the 

%interpretable models are important for various prediction tasks \citep{brenning2023interpreting}. By providing clear explanations for the detected stances, interpretable models provide insights into the decision-making process and allow users to understand why a particular attitude was recognized. This motivated us to develop a ZSSD module that is interpretable in its design and provides transparency, especially when classifying stances on sensitive social issues.
\par To achieve this, we first focus on deciphering the short snippets in the input post \citep{deyoung2019eraser}, which provide enough evidence to support the classification results and are referred to as \textit{implicit rationales}. Furthermore, it is possible that a single post contains different rationales that correspond to different attitudes toward the given target. For example: \textit{Target:} ``nuclear mission"; \textit{Text:} ``..I was against Nuclear power..but now it seems that nuclear should be in the mix. Fission technology is better..to be explored.."; \textit{Favor:} ``nuclear should be in the mix", ``Fission technology is better"; \textit{Against:} ``I was against Nuclear power". %Recently, large language models (LLMs) have been used for stance detection (SD) \citep{ding2024edda,zhang2024llm}. 
Hence, we focus on extracting all possible implicit rationales from the given input. Our preliminary investigations with LLMs motivated us to leverage LLMs to extract rationales rather than to identify stances and best-possible rationale for predictions (Section \ref{sec_res}). Moreover, due to cost and time constraints, it is not possible to achieve ground-truth stance labels for all rationales for all inputs.
%, and further develop an efficient architecture providing interpretability (Section \ref{sec_res}). 
%Due to the popularity of LLMs, we initially investigated the performance of LLMs for ZSSD in zero, few-shot, and fine-tuning settings and for the extraction of the best-possible rationale for the predicted attitude, but it resulted in the unsatisfactory performance of LLMs (Section \ref{sec_res}). Instead of capturing relevant rationales and predicting correct stances, LLMs prove to be effective in extracting all possible implicit rationales. In contrast to the existing works using LLMs for SD \cite{ding2024edda,zhang2024knowledge}, our preliminary investigation motivated us to leverage LLMs for extracting implicit rationales and further develop an efficient architecture providing interpretability. Moreover, due to cost and time constraints, it is not possible to achieve ground-truth stance labels for all rationales for all input posts. Therefore, in our work, we use ranking algorithms to interpret the relevance of each implicit rationale with respect to different stance labels, so that we do not have to rely on the ground truth of the rationales. %This minimizes human effort and automatically captures the context of each rationale with respect to different stances and targets. 
Therefore, we develop different stages of our model, namely the ``relevance ranking", which automatically interprets the relevance of each rationale to favor, against, and neutral stance using ranking algorithms, while the ``grouping and selection" stage selects the top-k most diverse relevant and irrelevant rationales, thus guiding the model to correctly predict stance with the reasoning for such prediction, thus ensuring inherent interpretability.\\
\indent In addition, the linguistic measures of communication dynamics, such as empathy, absolutism, action, abstract, concrete, communion language, etc., provided by the LIWC framework \citep{boyd2022development} could offer deeper insights into how and why a user adopts a certain attitude. For example, empathic or communicative language may indicate a supportive stance, while absolutist or action-oriented language could indicate strong opposition. For instance: \textit{Target:} ``prices"; \textit{Stance:} ``Against", \textit{Text:} ``...Increased..electricity..gas prices again..incompetent, stupid, useless..Pwehsident..! Leche!"; \textit{Linguistic assessment (LLM):} ``The post exhibits low empathy as it contains insulting language..confrontational and critical, indicating absolutist thinking..approach language reflects anger and frustration to the issues of rising price..no signs of allure or persuasion, instead, the language is direct and attacking.". Hence, these reasonings as \textit{explicit rationales} help to uncover hidden biases, motivations, or attitudes in a post to provide a more interpretable understanding of the user's attitude towards the target. We utilize the reasoning capabilities of LLMs to assess the input based on several linguistic features (refer Section \ref{sec_rat_gen}). The final concatenated relevant implicit and explicit rationales help to classify the final attitude of the given input. \\
%For example: \textit{Target:} "prices"; \textit{Stance:} "against", \textit{Text:} "...Increased..electricity..gas prices again..incompetent, stupid, useless..Pwehsident..! Leche!"; \textit{Linguistic assessment (LLM):} "The post exhibits low empathy as it contains insulting language..confrontational and critical, indicating absolutist thinking..approach language reflects anger and frustration to the issues of rising price..no signs of allure or persuasion, instead, the language is direct and attacking.". Hence, this motivated us to utilize linguistic measures as \textit{explicit rationales} to uncover hidden biases, motivations, or attitudes in a post along with the implicit rationales to provide a more interpretable understanding of the user's attitude towards the target. The final concatenated relevant implicit and explicit rationales help to classify the final attitude of the given input.\\
\indent The main contributions of our work are as follows: \\
\textbf{\textit{(i.)}} To the best of our knowledge, this is the first study that considers stance detection as an information retrieval ranking task while providing inherent interpretability by focusing on both implicit and explicit reasonings to uncover the hidden stance-related information in the post. \textbf{\textit{(ii.)}} We propose a novel interpretable ZSSD framework, consisting of 3 main stages, \textit{relevance ranking, grouping and selection, and classification} that leverages implicit rationales as a subsequence of words and explicit rationales based on linguistic measures to classify the stance. We refer to our model as \textbf{I}nterpretable \textbf{R}ationales for \textbf{S}tance Detection as \textbf{IRIS}.
\textbf{\textit{(iii.)}} In our proposed relevance ranking stage, stance detection is considered as a ranking task to automatically assign each implicit rationale towards stances, while the grouping and selection phase selects the k most diverse rationales. In this way, the disadvantage of the lack of human-annotated ground truth for rationale stances is overcome, leading the model towards relevant rationales and supporting interpretability by design. \textbf{\textit{(iv.)}} The classification phase of IRIS uses the encoded representations of the implicit and explicit arguments to predict the stance of each selected rationale and classifies the final stance of the inputs using majority voting. \textbf{\textit{(v.)}} Extensive experiments are conducted on VAST and EZ-STANCE datasets for ZSSD and P-Stance and RFD datasets for generalizability analysis using $\textit{50\%}$, $\textit{30\%}$, and even $\textit{10\%}$ training data to demonstrate the efficiency of our IRIS. %The code and datasets are present here: \url{https://osf.io/ja97s/?view_only=bb6402c6ddce4bc482deeb879a0096d5} (please make sure to add `\_' after view and remove space after `=' in URL.)

%{\footnote{https://osf.io/ja97s/?view\_only=bb6402c6ddce4bc482deeb879a0096d5}}. 

\par  
\begin{comment}

\begin{figure}
\centering
\includegraphics[width=0.40\linewidth]{flow_vertical.png}
 \caption{Flow diagram of our proposed approach IRIS.}
  \label{fig_flow_vert}
\end{figure}
\end{comment}
\section{Related Work}
\subsection{Stance Detection}
A wealth of work has been explored in the field of in-target and cross-target SD, where training and test targets are the same or closely related \citep{ upadhyaya2023toxicity}. However, recognizing the stance of an unknown target appears to be of greater importance as training data is not available for all targets. Therefore, recent literature aims to address the ZSSD. Recent works on ZSSD have either focused on target-specific and invariant features \citep{zhang2024commonsense} or used contrastive learning to improve stance detection in zero-shot scenarios \citep{yao2024enhancing}. Building on advancements in prompt learning, \citet{yao2024enhancing} integrated prompt learning with contrastive learning to improve zero-shot stance detection. However, \citet{wang2024meta} introduced a meta-learning algorithm combined with data augmentation to address generalizability challenges associated with stance detection. Additionally, \citet{zhao2024zero} proposed a collaborative feature learning framework with multiple experts for zero-shot stance detection. Despite these innovations, these methods continue to face limitations in performance on unseen targets, often struggling to capture nuanced stance-related information and lacking transparency or contextual understanding between targets and text. This motivated us to design a framework capable of uncovering the underlying reasoning and motivations behind the author's attitude toward a given target. By extracting both implicit and explicit reasons, our proposed approach provides a clear explanation for the stance expressed in the post.

\begin{figure*}
\centering
\includegraphics[width=0.75\linewidth]{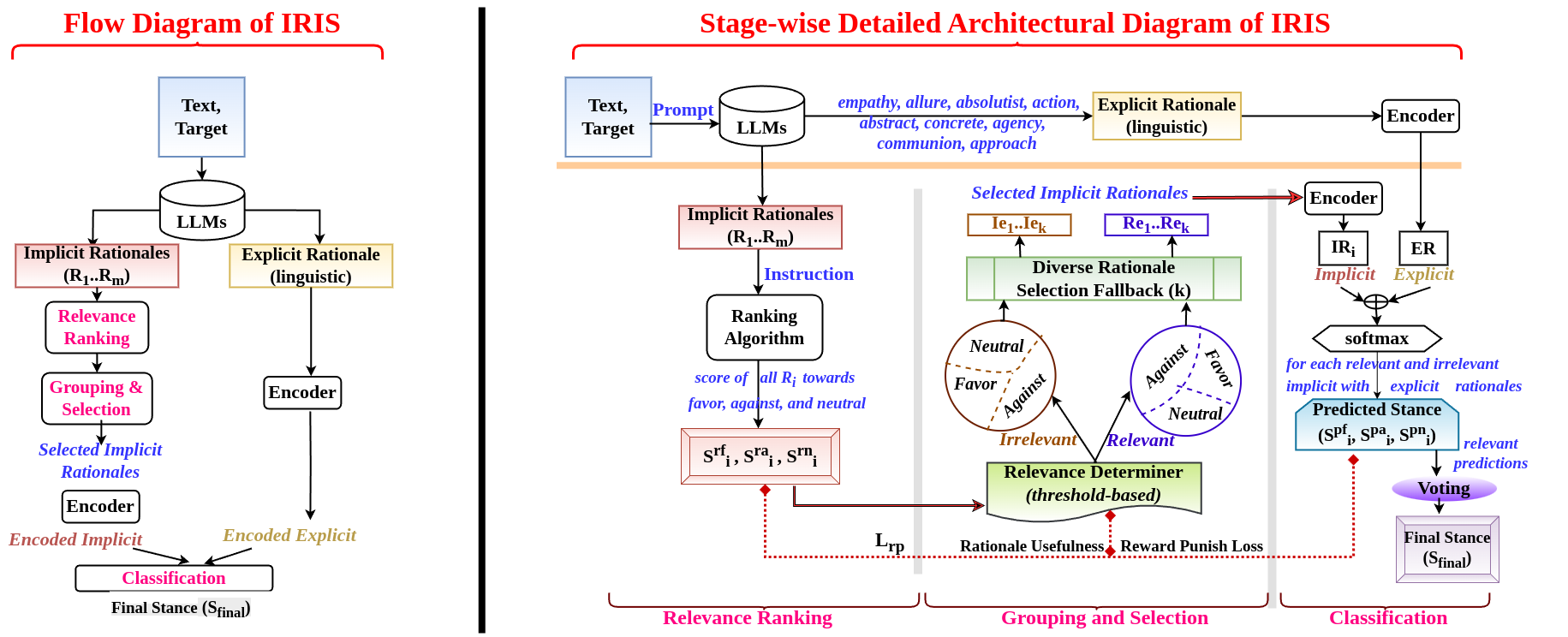}
 \caption{Flow diagram of IRIS followed by stage-wise detailed architecture design. In stage-wise diagram: \textit{[Left]: Relevance Ranking; [Center]: Grouping and Selection; [Right]: Classification}. Notations: $R_i$: LLM-generated $i^{th}$ implicit rationale; $\{S^{rf}_i, S^{ra}_i, S^{rn}_i\}$: Relevance scores of $i^{th}$ implicit rationale towards favor, against, and neutral stances; ${Re}_k$, ${Ie}_k$: k relevant and irrelevant implicit rationales; ${IR}_i$: Encoded $i^{th}$ relevant implicit rationale; ${ER}$: Encoded explicit linguistic rationale; $\{S^{pf}_i, S^{pa}_i, S^{pn}_i\}$: stance scores for each encoded relevant $i^{th}$ implicit rationale and explicit rationale;
$S_{final}$: Final predicted stance. The details are clearly mentioned in Section \ref{sec_methodo}.}
  \label{fig_flow}
\end{figure*}

\subsection{LLMs for Stance Detection}
Recent trends are moving towards the use of LLMs for ZSSD. \cite{zhang2024knowledge} used LLMs to bridge the gap between seen and unseen targets and effectively integrate this transferable knowledge to guide stance predictions. \cite{zhang2024llm} utilized LLM to explicitly extract the relationship between paired texts and targets, while \cite{hu2024ladder} extracted target information from the web to improve the attitude task. The chain-of-thought context learning is exploited by \cite{taranukhin2024stance}, which applied explicit reasoning using LLM to classify stance. Moreover, some of the more recent methods are directed towards explanations for stance detection \citep{zhang2024knowledge}. \cite {ma2024chain} created a prompting framework that models stance detection as a sequence of stance-related assertions. \cite{ding2024distantly} employed the instruction-based chain-of-thought (CoT) method to extract stance recognition explanations from a very large language model. In addition, \cite{saha2024stance} focused on generating explanations for predicted stances by capturing the pivotal argumentative structure embedded in a document. However, our approach formulates stance detection as a ranking task that aims to inherently capture the relevance of implicit rationales for different stances while using communicative features to decode the emotional and cognitive dimensions of the attitude. This eliminates the need for ground truth for rationales while providing a more comprehensive understanding of the author's attitude. Even recent work has highlighted the challenges of interpretability in the era of LLMs \citep{singh2024rethinking}. Hen, our work offers advantages over existing works in terms of interpretability, rationale analysis, and considering specific text subsequences in relation to stance and linguistic cues.

%This motivated us to develop an interpretable approach that focuses on extracting the relevant snippets/implicit rationales, thus leading the model to predict the actual attitude while utilizing the linguistic features of the given input to decode the entire stance.

\section{Methodology \label{sec_methodo}}
%\textcolor{red}{The Methodology section is quite comprehensive, but to improve readability and clarity, I would suggest breaking down complex sentences into smaller, more digestible parts. Here's a general suggestion: Clarify Motivation Early: Start each subsection by explicitly stating why each step or compnoent is necessary and how it ties into the overall goal of ZSSD (Zero-Shot Stance Detection). Example:At the beginning of each stage, introduce a brief one-liner explaining the core need or motivation behind it. For instance, for Rationale Generation, mention upfront that the goal is to ensure that both explicit and implicit reasoning behind stance decisions is captured for interpetability.}
\textbf{Problem Definition:} Given a training dataset containing $P$ samples with text, target, and the corresponding stance towards the target (${(x_i,t_i,s_i)}_{i=1}^{P}$) and a set of $Q$ samples as a test set with text and targets that cannot be seen during training (${(x_i,t_i)}_{i=1}^{Q}$). The aim is to design an interpretable ZSSD to determine the stance of the unseen test set (favor/against/neutral), achieved by inherently focusing on relevant implicit rationales together with the explicit reasonings without the human annotations (ground truth) for rationales.
\par \noindent Figure \ref{fig_flow} represents the flow of our proposed approach IRIS, which consists of 3 main stages of \textit{Relevance Ranking, Grouping and Selection, and Classification}. The input post is first fed into the LLM to extract the explicit rationales based on linguistic features and the implicit rationales, i.e. the subsequence of words from the input text that lead to the different stance categories towards the given target. The implicit rationales are then fed to the \textit{relevance ranking} stage which determines the ranking of the implicit rationales for a favorable, against, or neutral attitude. %This stage consists of 3 ranking mechanisms (a pre-trained ranker, our proposed ranker, and LLM-based scores) that results in the final the weighted score for each extracted rationale towards favor, against, and neutral stance. 
Each rationale is then categorized into relevant or irrelevant groups and the k most informative and diverse rationales are then selected in the \textit{grouping and selection} phase and encoded using sentence transformer. %Our IRIS model then selects  from the relevant and irrelevant groups that reflect different viewpoints. 
In parallel, the explicit rationales are also encoded separately to capture meaningful context. The selected implicit rationales together with the explicit reasoning then pass through the softmax layer to determine the stance for all rationales. To classify the final stance of the input, the attitudes predicted by the relevant rationales are guided by the consensus and result in the final stance label. Next, we describe the different stages in detail.

\subsection{Rationale Generation}\label{sec_rat_gen} 
This stage ensures that both explicit and implicit reasoning behind stance decisions is captured for interpretability. Here, the input text with the target is fed into the LLM to extract rationales. As the same input text can have different stances depending on the target, we apprehend the target with the input text. %The linguistic measures of communication dynamics, such as empathy, absolutism, etc. provided by LIWC framework \citep{boyd2022development} help infer the emotional and cognitive dimensions of the stance by understanding how and why the author takes a certain stance. 
Using the prompt given in Figure \ref{fig_prompt_ex}, we ask the LLM to assess the given post based on the linguistic measures and provide a more interpretable understanding of the user's attitude towards the target, by viewing them as explicit rationales. To ensure inherent interpretability in stance task, we use the prompt (Figure \ref{fig_prompt_im}) to extract the subsequence of words associated with different attitudes as a set of implicit rationales. In the following, we explain the components that help in processing the implicit and explicit rationales.

\subsection{Relevance Ranking \label{sec_rank}} %Stage is responsible for providing transparency by extracting relevance of ratonales
The main responsibility of this stage is to provide transparency by designing a ranking algorithm that inherently interprets the relevance of each rationale towards different stance labels as we do not rely on ground truth for rationales (Figure  \ref{fig_flow}[Left]). This helps in minimizing human efforts and automatically captures the context of each rationale towards different stances and targets, thus directly contributing to our model's decision-making process. Here, we use the LLM extracted implicit rationales retrieved from the previous \textit{rationale generation} stage (Section \ref{sec_rat_gen}) as a set of m rationales $\{R_1, R_2,...R_m\}$ (``m" may vary for each input depending on the LLM responses). We consider each \textit{rationale} with the \textit{target} as a \textit{query} and prepare a set of $3$ documents containing various posts and their targets corresponding to ``favor", ``against", and ``neutral" stances without explicitly mentioning the stances in the documents.
\subsubsection{Document Preparation} The documents are articulated using the publicly available benchmark zero-shot stance datasets. To ensure novelty and minimize overlap, we only include statements in the documents that have a cosine similarity of less than $0.05$ with training and test data. This filtering process prevents the use of information that is similar to the samples used for model evaluation, thus reducing the risk of data leakage and maintaining the integrity of the zero-shot framework. We intentionally exclude any stance labels from the documents, ensuring that there is no bias introduced into the training process. The documents may contain statements that align with favor, against, or neutral perspectives, but these are kept implicit and without stance labels to avoid any task-specific bias. Hence, these documents act as a source of external knowledge for the ranker to compute relevance scores that indicate how closely a given implicit rationale as a query aligns with the content of each document. By excluding stance labels and ensuring novelty through cosine similarity filtering, the documents preserve the zero-shot nature of the stance detection task.
\begin{figure}
\centering
\includegraphics[width=0.55\linewidth]{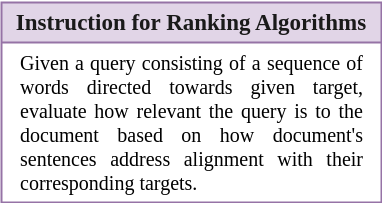}
 \caption{Instruction for Ranking Algorithm}
  \label{fig_instruct}
\end{figure}

\subsubsection{Instruction}
In addition to the query and set of documents, we first construct an instruction for the ranker (refer Figure \ref{fig_instruct}).
% \textit{"Given a query consisting of a sequence of words directed towards given target, evaluate how relevant the query is to the document based on how document's sentences address stance alignment with their corresponding targets."}. 
This helps the rankers understand contextual relationships between document-target connections and allows the ranker to infer relevance without revealing explicit stance labels. The query-document pair together with the instruction is then fed to the ranker.

%We ensure to keep the vocabulary and information within the documents as external knowledge. 
%We do not consider explicit stance labels in the documents, thus avoiding the potential leakage of any data during the training of the models. We utilize only those samples from public stance datasets

%We then experiment with LLM and other ranking algorithms to compute the relevance of each extracted rationale towards different stances (refer Section \ref{}). We observed that 

%However, our initial experiments suggest that LLMs did not perform best in detecting the stances and  (refer Section \ref{sec_res}), therefore to avoid over-reliance on LLM's scores and build a more trustworthy system, we also use other pre-trained and our developed ranking algorithms. While LLM is being explicitly prompted to generate probabilities related to stances of rationales towards targets, the rankers compute relevance in a more implicit way based on query-document matching mechanisms. 
\subsubsection{Ranking Algorithm} We experiment with different pre-trained rankers and LLM, as described in Section \ref{sec_res}. We then select the best ranking algorithm for our study as FlagReranker \cite{xiao2023c}. Inspired by \cite{chen2024fintextqa}, we also employ a publicly available bgereranker\footnote{https://huggingface.co/BAAI/bge-reranker-large} to obtain the relevance scores of each rationale towards favor, against and neutral documents. The ranker provides a list of $3$ relevance scores which are often raw and unbounded, therefore passed to the softmax function to obtain a probabilistic distribution, as probabilities provide a clearer understanding of the relative confidence for each stance, compared to raw values, resulting in $\{S^{rf}_i, S^{ra}_i, S^{rn}_i\}$ for all $R_i$. We utilize the well-established pre-trained FlagReranker to ensure that scores are grounded in the underlying semantic relationships. By identifying and ranking rationales, the system inherently highlights key supporting or opposing evidence, making the stance decision more explainable. Thus, the use of ranker adds a layer of inherent interpretability to the stance detection pipeline.

\subsection{Grouping and Selection} 
\label{sec_group}
%The main task of this stage is to divide implicit rationales into relevant and irrelevant groups and output the top-k diverse rationales for further processing. 
This stage identifies relevant and irrelevant rationales towards stances (Figure  \ref{fig_flow}[Center]). This results in improving interpretability by clearly indicating both the positive (relevant) and negative (irrelevant) influences on the stance, making the model's final predictions more explainable.
\subsubsection{Relevance Determiner} We use the threshold-based difference to determine whether a rationale is relevant to a particular stance and irrelevant to others.
%to measure the relative importance of a rationale across different attitudes. 
%Since we have scores for each document, the relevance decision is based on how much the rationale's score for one document ($d_f, d_a, d_n$) differs from the others. 
Given a relevance score of the rationale ($R_i$) $S^{rf}_{i}, S^{ra}_{i}, S^{rn}_{i}$ obtained from the previous ranking stage, we determine if $(S^{rf}_{i} - max(S^{ra}_{i}, S^{rn}_{i})) > threshold$, then consider the rationale as relevant for the favor stance and irrelevant for against and neutral stance; a similar process is repeated for all stance labels of all rationales. We also consider other approaches for grouping that use clustering methods, however, we already have relevance scores from the previous phase. Moreover, our component provides an unsupervised approach with transparency to identify relevant rationale based on score distributions without requiring ground truth labeling, which led us to utilize this method in determining the relevance of all generated rationales. The output of ``Relevance Determiner" results in 2 groups of relevant and irrelevant rationales consisting of favor, against, and neutral stances as sub-groups. 
\subsubsection{Diverse Rationale Selection Fallback (k)} aims to select diverse rationales from two main groups (relevant and irrelevant) while balancing the representation of subgroups within each group (favor, against, and neutral). %The challenge is to avoid over- or under-representation of a subgroup and instead aim for a balanced selection. 
In a real environment, it is often impossible to have the same number of rationales for each subgroup (favor, against, neutral). To achieve this, we use KL-divergence to minimize the deviation from a dynamic target desired distribution based on available rationales. The detailed steps are explained in Algorithm \ref{alg_div1}. The algorithm starts by computing a target distribution $V$ based on the proportion of rationales available in each subgroup (steps 1-3 of Algorithm \ref{alg_div1}). Using KL-divergence, the algorithm iteratively selects rationales to minimize the divergence between the current distribution of selected rationales and the target distribution. Initially, no rationales are selected, so the current distribution is initialized with small non-zero values to avoid computation errors (step 4 of Algorithm \ref{alg_div1}). At each step, a rationale is temporarily added, the new distribution is calculated, and the rationale that best aligns the updated distribution with the target is chosen (steps 8-13 of Algorithm \ref{alg_div1}). The process continues until the desired number ($k$) of rationales is selected. A fallback mechanism ensures progress even if one subgroup is exhausted by selecting from the remaining groups (step 15 of Algorithm \ref{alg_div1}). This process is applied separately for relevant and irrelevant implicit rationales to maintain balanced coverage.

%This approach dynamically adapts to the availability of rationales in each subgroup, ensures no errors or over/under-representation, and selects rationales in such cases also where diversity cannot be perfectly preserved due to missing subgroups, providing a fallout mechanism.
\par This stage finally leads to $k$ relevant (${Re}_1..{Re}_k$) and irrelevant (${Ie}_1..{Ie}_k$) selected implicit rationales.

\subsection{Classification}
Here, the main aim is to classify input's stance (Figure \ref{fig_flow}[Right]). Loss functions improve the IRIS performance and emphasize clarity in the decision-making of predictions. 
\subsubsection{Rationale Encodings} As the previous grouping stage results in implicit rationales $k$ relevant (${Re}_1..{Re}_k$) and irrelevant (${Ie}_1..{Ie}_k$) in a given context, these implicit rationales, together with their input targets, are then passed through the sentence encoder \cite{{lee2024nv}} of the embedding dimension ($d_e$) followed by a dense layer ($d_d$), resulting in $IR_i \in \mathbb{R} ^ {1 \times d_d}$, helping to understand the meaningful context within the rationales. These embeddings have been effective in various retrieval and classification tasks \cite{yang2024diagnosing, lee2024nv} (different embeddings have been examined in Section \ref{sec_res}). In parallel, the explicit rationales extracted by LLM are separately encoded using sentence embeddings ($d_e$) \cite{{lee2024nv}} and passed through a dense layer ($d_d$). This captures the essence of the context between the input and the linguistic characteristics, leading to $ER \in \mathbb{R} ^ {1 \times d_d}$. 
\subsubsection{Stance Prediction} For each selected implicit rationale, the implicit and explicit encoded rationales are concatenated (${R} ^ {1 \times 2(d_d)}$) and passed through softmax, resulting in the stance prediction of each rationale towards favor, against, or neutral stance $\{S^{pf}_i, S^{pa}_i, S^{pn}_i\}$. We select the stance predictions of concatenated explicit and relevant implicit rationales and classify the final stance ($S_{final}$) of the input based on the majority vote. If no decision is made, we classify the output as neutral. Various loss functions are as follows.
\subsubsection{Loss functions} 
\textit{\textbf{Stance Loss ($L^{s}_{{ce}}$):}} The categorical cross-entropy ($ce$) loss is calculated for the final predicted stance of input $S_{final}$ and the true stance label for ZSSD. \\
%\textit{\textbf{Weighted Ranking Judgement Loss ($L_{rj}$):}} is introduced to test the effectiveness of our rankers used in the relevance ranking phase. Since we do not have human-labeled ground truth for the stance of each rationale, we consider the relevance ratings as a proxy for the ground truth to determine how closely the predicted viewpoints match the ranking-based truth. We use stance labels by LLM, FlagReranker, and CARR as ground truth and $\{S^{pf}_i, S^{pa}_i, S^{pn}_i\}$ as predicted stance labels. The weighted loss helps to improve the overall performance of our model by offering insight into how our final stance predictions are relative to the rankers, thus providing insight into the rankers' weaknesses. $L_{rj}= w_1L^{l}_{ce}+ w_2L^{r}_{ce} + w_3L^{c}_{ce}$. \\
%\textit{\textbf{L2 Regularization:}} penalize higher weights ($w_1, w_2, w_3$) assigned to our CARR algorithm, FlagReranker, and LLM respectively to reduce overfitting and prevent a single weight from dominating.\\
\textit{\textbf{Rationale Usefulness Reward Punish Loss ($L_{rp}$):}} This custom loss calculates how much the relevance ranking phase should reward/penalize depending on whether or not the rationales selected resulted in the correct stance in the final stage. It is based on the idea that the selection of relevant rationales should lead to correct stances, while irrelevant rationales should result in incorrect predictions. We focus on the loss in the relevance phase ($L^{rel}_{ce}$), where $S^{rf}_{i}, S^{ra}_{i}, S^{rn}_{i}$ are prediction scores versus ground truth stance. We also examine the relevance labels of the rationales (relevant/irrelevant) based on the grouping stage. $L_{rp} = L^{rel}_{ce} \times (1-R)$, where R is a reward/punishment term defined using the stance predictions of the classification stage as follows: $R= \beta (reward)$: if relevant rationale leads to correct stance or irrelevant rationale leads to incorrect stance predictions; while $R= -\beta  (punish)$: if relevant rationale results in incorrect stance or irrelevant rationale leads to correct predictions [here the predictions are $\{S^{pf}_i, S^{pa}_i, S^{pn}_i\}$ of classification stage compared to ground truth stance] (refer Figure \ref{fig_flow}). This adjustment encourages the relevance ranking stage to improve its rationale selection strategy based on stance prediction outcomes in the classification stage.\\ %\textit{\textbf{Adversarial Contrastive Loss ($L_{ac}$):}} works in adversary to the relevant and irrelevant rationale embeddings on the basis of classification stage stance predictions of rationales $\{S^{pf}_i, S^{pa}_i, S^{pn}_i\}$ versus ground-truth stance. If the prediction is correct, we try to minimize the distance between the relevant and irrelevant embeddings, working against the model to better separate relevant and irrelevant rationales. If the prediction is incorrect, we apply a margin-based penalty to promote the distance between rationale embeddings to help the model distinguish between relevant and irrelevant embeddings. This loss is achieved by $L_{ac}= \frac{1}{k.k}\sum_{i=1}^{k}\sum_{j=1}^{k} y_iD(E^R_i, E^I_j) + (1-y_i).max(0, margin-D(E^R_i, E^I_j))$, where $y_i=1$ if the prediction is true, 0 otherwise, $D$ is the euclidean distance, $k$ is the no. of selected rationales, $E^R$ and $E^I$ are embeddings of relevant and irrelevant rationales . 
{\textbf{\underline{Total loss:}}} of our proposed approach is, $L = L^{s}_{{ce}} + qL_{rp}$, where q is ratio of rationale usefulness loss to total loss.

\begin{table}
\centering
\scalebox{0.85}{
\begin{tabular}{|l|l|l|l|}
\hline
\textbf{} & \textbf{Train} &\textbf{Dev} &\textbf{Test} \\ \hline
\# Examples &13477 &2062 &3006 \\ 
\# Unique Comments &1845 &682 &786 \\ \hline
\# Zero-shot Topics &4003 &383 &600 \\ 
\# Few-shot Topics &638 &114 &159 \\ \hline
\end{tabular}
}
\caption{Dataset statistics of VAST}
\label{tab_data_vast}

\end{table}

\begin{table}
\centering
\scalebox{0.85}{
\begin{tabular}{|l|l|l|l|}
\hline
\textbf{} & \textbf{Train} &\textbf{Val} &\textbf{Test} \\ \hline
\# Samples of noun-phrase targets &13756 & 2354 &2663 \\ \hline
\# Samples of claim targets &18879 & 4349 &5135 \\
\hline
\end{tabular}
}
\caption{Dataset statistics of EZ-STANCE}
\label{tab_data_ez}
\end{table}
\section{Experimental Setup \label{sec_setup}}
\subsection{Datasets \label{sec_dataset}}

To evaluate our IRIS model, we conduct experiments on two \textbf{zero-shot} stance detection (ZSSD) benchmarks: VAST and the recently curated EZ-STANCE. VAST comprises comments from The New York Times’ Room for Debate section, while EZ-STANCE contains tweets collected from social media discourse. These datasets together offer a rich variety of targets, including both noun phrases and claim-based target-text pairs, thus enabling a broad and comprehensive evaluation of model performance across informal and structured contexts. \textit{\textbf{VAST \cite{DBLP:conf/emnlp/AllawayM20}:}} ZSSD dataset with \textit{pro, con, neutral} labels for both zero-shot and few-shot targets in the train and test set. The dataset statistics are present in Table \ref{tab_data_vast}. \textit{\textbf{EZ-STANCE (EZ) \cite{zhao2024ez}:}} recently curated tweets including noun-phrase (N), claim-based (C), and mixed (M) targets with \textit{Favor, Against, Neutral} stance labels. The statistics are available in Table \ref{tab_data_ez}.  

\subsubsection{Generalizability Analysis:} To further assess IRIS’s robustness across domains, stance types, and linguistic styles, we extend our evaluation to two additional datasets: P-Stance and the RFD News Articles dataset. P-Stance focuses on political figures, enabling domain-specific generalization analysis within political discourse. Importantly, we evaluate IRIS on both \textbf{in-target} and \textbf{zero-shot} stance settings for P-Stance, thereby testing its adaptability to both seen and unseen targets. RFD, on the other hand, provides long-form opinion articles from The New York Times’ Room for Debate section—similar in source to VAST but distinct in content length and writing style. While VAST samples average around 100 words, RFD articles average 416 words, the highest among all datasets (EZ-STANCE: 40; P-Stance: 30). This extended length introduces greater discourse complexity and nuanced stance expression, making RFD particularly valuable for evaluating IRIS’s generalizability under \textbf{in-target} settings in formal, context-rich environments. \textit{\textbf{{P-Stance \cite{li2021p}:}}} is a political stance dataset containing $7,953$ annotated tweets for “Donald Trump”, $7,296$ for “Joe Biden”, and 6,325 for “Bernie Sanders” as target domains with favor and against stances. \textit{\textbf{RFD \cite{saha2024stance}:}} consists of a total of $764$ claim-based article pairs with pro ($44\%$), con ($47\%$), and balanced ($9\%$) stance labels. Despite its small size, it has the highest article/text length (416 average words) when compared with all other datasets, as mentioned above (Average number of words- VAST: 100; EZ-STANCE: 40; P-Stance: 30).

All datasets consist of English-language posts and do not include any personally identifiable information; however, they may contain offensive content due to their explicit stances on topics such as religion, politics, immigrants, etc. We strictly adhere to the requirements of the respective licenses for all datasets used in our study. \textit{\textbf{Please note that VAST and EZ-STANCE serve as the primary datasets for evaluating ZSSD in our study, while P-Stance and RFD are utilized to assess the generalizability of our approach across domains and settings.}}
\begin{figure}
\centering
\includegraphics[width=0.85\linewidth]{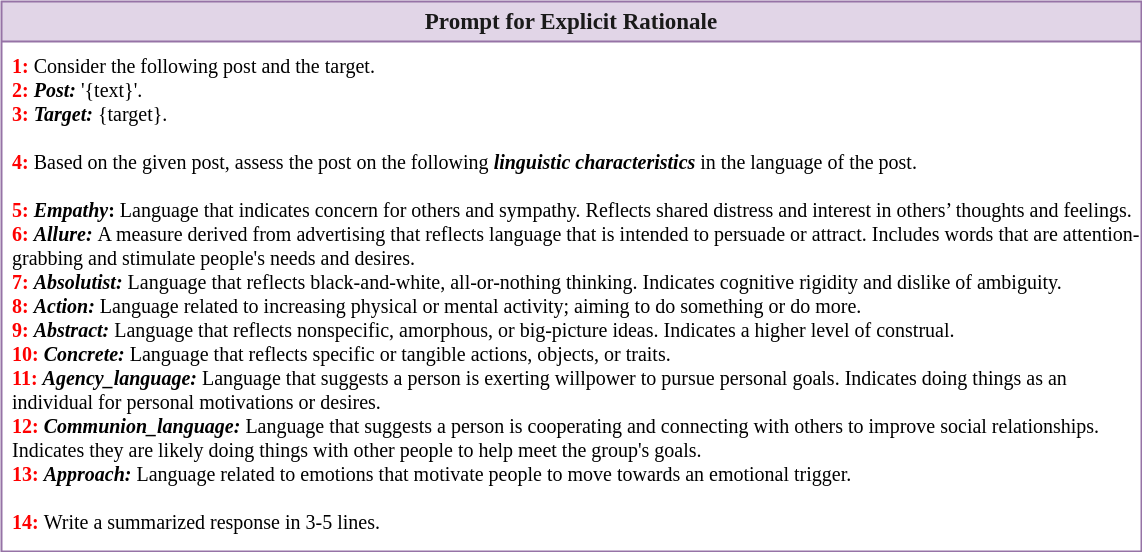}
 \caption{LLM Prompt for Explicit Rationale}
  \label{fig_prompt_ex}
\end{figure}

\begin{figure}
\centering
\includegraphics[width=0.85\linewidth]{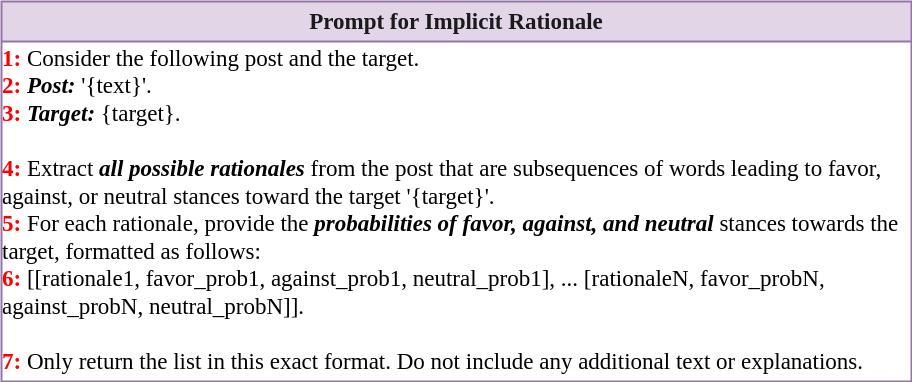}
 \caption{LLM Prompt for Implicit Rationale}
  \label{fig_prompt_im}
\end{figure}

\subsection{Rationale Generation and Annotation for LLM-Based ZSSD}

\subsubsection{LLM Prompts for Rationale Generation \label{sec_llm_prompts}}
A pilot study was conducted on a randomly selected small sample of 100 posts from the VAST and EZ datasets to assess the efficacy of our prompts. This exploratory analysis enabled us to evaluate prompt responsiveness in a controlled setting, informing subsequent refinements that optimized the prompts utilized in our experimental design. Figures \ref{fig_prompt_ex} and \ref{fig_prompt_im} illustrate the final prompts we used to ask Llama 3.1 to generate explicit and implicit rationales in zero-shot settings respectively. As mentioned in Section \ref{sec_res}, Llama 3.1 performs better than Mistral on the ZSSD task. We take advantage of the text generation capabilities to use Llama for rationales generation. For the case of \textit{explicit rationales} (Figure \ref{fig_prompt_ex}), we derive definitions for various linguistic measures from LIWC \cite{boyd2022development} and Receptiviti’s LIWC framework using the `liwc\_extension' object\footnote{\url{https://docs.receptiviti.com/frameworks/liwc-extension} \label{liwcurl}}. This allows us to extract scores for the following attributes: empathy, allure, absolutist, action, concrete, agency\_language, communion\_language, and approach. For full definitions, please refer to the documentation\footref{liwcurl} and Figure \ref{fig_prompt_ex} (Steps 5-13). We omit the avoidance measure as it is simply the inverse of approach emotions and does not provide additional linguistic value. These linguistic measures focused on understanding the dynamics of communication and the determinants of interpersonal support, thus helping us to decipher the reason for the stance of the text. We then prompted Llama 3.1 to assess the following linguistic characteristics based on the given textual post and target (Steps 1-4 of Figure \ref{fig_prompt_ex}) and generate a summarized response in 3-5 lines (Step 14 of Figure \ref{fig_prompt_ex}).\\
\indent To understand the relevance of each rationale for the different viewpoints, we use the prompt mentioned in Figure \ref{fig_prompt_im} to extract the all possible implicit rationales based on given input post and target (Steps 1-4 of Prompt \ref{fig_prompt_im}) and retrieve their relevance scores towards favor, against, and neutral stances (Step 5 of Prompt \ref{fig_prompt_im}). Please note that we use these probabilistic scores directly in the relevance ranking phase of our IRIS as part of the ablation study to investigate LLM scores being used to determine the relevance of rationales as described in Section \ref{sec_res_ranking}. We manually analyzed the LLM responses for extracting implicit and explicit rationales for 100 randomly selected posts. A team of two PhD students and one postdoctoral researcher with a background in linguistics and computer science were assigned to examine the LLM responses and were given clear instructions for the task. The annotators observed satisfactory and convincing results, therefore, we used LLMs in the rationale generation stage of our IRIS. 

\subsubsection{Rationale Annotation for LLM's Correct and Incorrect Predictions for ZSSD} \label{app_rationale}
As mentioned in Section \ref{sec_res}, to understand the interpretability provided by LLMs for zero-shot stance task, we investigate the random $100$ correct and incorrect predictions of Llama 3.1 and prompt the LLM to extract the best possible subsequence of words from the input text leading to the predicted stance by LLM. In order to evaluate the LLM's performance on implicit rationale extraction, we manually annotate those 100 samples, considering them as ground truth. The annotation team consists of three annotators who are proficient in English and are doctoral and post-doctoral researchers in the field of linguistics and computer science. They have a strong background and in-depth knowledge of natural language processing,  computational social sciences, and information retrieval tasks, which predestines them for this task. We consider rationale identification as a binary classification task. The aim is to give each input token a binary label, suggesting whether the token is present in the rationale or not. These clear instructions were given to the annotators to manually annotate the ground truth for selected $100$ samples. We then compared the ground truth versus the Llama responses for the rationale and observed a significantly lower F1 score of $0.617$, which motivated us to find other better ranking algorithms that can be beneficial in providing relevance scores for each rationale towards different stances (Section \ref{sec_perf_llm}). To further assess the quality of implicit and explicit rationales by IRIS for ZSSD, we conduct both automatic and human evaluations in Section \ref{sec_eval_rat}.

\subsection{Implementation Details} \label{sec_impl}
%We use the publicly available training and test sets of all datasets. For the EZ dataset, we focus on subtask A, which involves identifying target-based ZSSD, including noun-phrase, claim-based, and mixed targets. As mentioned in Section \ref{sec_dataset}, we use VAST and EZ-Stance for ZSSD, while to prove the generalizability of our proposed approach in different domains and task type, we use P-Stance (in-target and zero-shot stance) and RFD (in-target stance) datasets. 
%However, to further prove the efficacy of our IRIS, 
%For P-Stance, we focus on both in-target stance where training and testing is done on same targets (for instance: for Trump target: training and testing is done on Trump tweets) and also for zero-shot stance task where  To determine the efficacy of our approach, we use $\textbf{10\%}$, $\textbf{30\%}$, and $\textbf{50\%}$ train samples for all used datasets.  For the VAST and EZ, we have 3 stance labels while for the P-stance and RFD, as mentioned in the proposed work \citep{li2021p,saha2024stance}, two stance labels (favor/against and pro/con) are used for training and evaluation.

%When we evaluate and train the model for the VAST dataset, we make use of samples from EZ as documents for the ranking stage (with cosine similarity $<$ $0.05$ and no stance labels). Similarly, we use zero-shot targets from VAST as documents for EZ. \textit{Please refer to Section \ref{sec_rank} for detailed knowledge of articulating documents to preserve zero-shot framework.} 

We use the publicly available training and test splits for all datasets. For EZ, we focus on Subtask A, which involves identifying target-based ZSSD across noun-phrase, claim-based, and mixed targets. Following previous work \cite{zhao2024ez}, we also train IRIS under $3$ settings on EZ: {(1)} using train set with mixed targets, {(2)} using only noun-phrase targets, and {(3)} using only claim targets. IRIS is then evaluated on corresponding test set: (1) mixed-targets, (2) noun-phrases only, and (3) claims only (refer Table \ref{tab_data_ez} for this setup). As described in Section \ref{sec_dataset}, VAST and EZ are used for evaluating ZSSD, while P-Stance and RFD are included to demonstrate the generalizability of IRIS model across different domains and task types. Specifically, P-Stance covers both in-target and zero-shot settings within the political domain, while RFD focuses on in-target stance detection in long-form news articles. To assess the efficacy of IRIS under limited supervision, we train using $\textbf{10\%}$, $\textbf{30\%}$, and $\textbf{50\%}$ of the available training data for each dataset. VAST and EZ datasets use $3$ stance labels, whereas P-Stance and RFD use the binary stance setup (favor/against and pro/con) for experiments, consistent with their original papers \citep{li2021p, saha2024stance}.
\par \noindent \textbf{Hyperparameters:} Embedding dimension ($d_e$): 4096; dense layer dimension ($d_d$) [with ReLu activation]: 128; threshold \textit{(Relevance Determiner)}: 0.3; $k$ \textit{(Diverse Rationale Selection)}: 3; $R$ ($\beta$) [reward/punish metric]: 0.1; output neurons [softmax activation]: 3 [VAST and EZ]/ 2 [P-Stance and RFD]; optimizer: Adam (0.0001 learning rate), batch\_size: 32. The best parameter values are selected using TPE in Hyperopt \cite{bergstra2013making}, which minimizes loss functions. We fine-tune the loss weight using Grid Search from Scikit-learn (\textit{q}=0.5 ($L_{rp}$)).

\par \noindent \textbf{Evaluation Metrics:} We perform 5 independent runs of our IRIS to account for variability and report average metric scores and standard deviation. Macro-F1 scores are used for VAST, EZ, and RFD datasets as proposed in their work \cite{DBLP:conf/emnlp/AllawayM20,zhao2024ez, saha2024stance}. Following previous work \citep{li2021p,upadhyaya2023toxicity}, we calculate $F_{avg}$, which represents the average of F1 scores for \textit{Favor} and \textit{Against} for the P-Stance dataset. 

\par \noindent \textbf{Implementation of LLMs for ZSSD:} Since Unsloth’s quantized models lead to lower GPU VRAM consumption and have been used in recent research \cite{kumar2024overriding}, we also work with Unsloth’s pre-quantized 4bit Llama-3.1-8B-Instruct\footnote{unsloth/Meta-Llama-3.1-8B-Instruct-bnb-4bit} and 4bit mistral-7b-instruct-v0.3\footnote{unsloth/mistral-7b-instruct-v0.3-bnb-4bit} models for zero-shot, few-shot, and fine-tuning of the LLMs for the stance task. For fine-tuning the LLMs, the specified hyperparameters resulted in the best fine-tuning performance using LoRA: rank of 8, alpha of 16, and dropout rate of 0.1. We employ the better-performing Llama 3.1 model on the stance task (refer Table \ref{tab_llm}) to extract explicit and implicit rationales by using prompts given in Section \ref{sec_llm_prompts}. 
\par \noindent \textbf{Environment Details:} GPU Model: NVIDIA A100 GPU servers with carbon efficiency of 0.42 kgCO$_2$eq/kWh, Library Version: tensorflow 2.12.0, torch 2.4.0+cu121, transformers 4.43.2. The training times for various IRIS variants and baseline models are provided in Appendix \ref{app_train_time}. %On average, it took approximately 35 minutes to train our IRIS with $50\%$ training data, excluding the extraction of rationales and relevance scores from LLMs and obtaining relevance scores of the extracted rationales from the pre-trained FlagReranker algorithm. The rationales and scores were obtained as part of the pre-processing before the execution of the IRIS. 

\subsection{Baselines \label{sec_baseline}}

We adopt recent state-of-the-art methods as baselines, focusing on those that report results across the datasets used in our study. To ensure a statistically fair comparison with our IRIS model, we re-run each reproducible baseline over five independent runs using our GPU setup (described in Section \ref{sec_impl}) and report the average performance metrics. However, for the VAST and RFD datasets, some baselines could not be reproduced due to the lack of publicly available code or incomplete details regarding hyperparameters and implementation. In such cases, we report their results as stated in the respective original publications without re-implementation.
\par \noindent {\textbf{VAST:}} We re-run the following baselines for five rounds: COLA \citep{lan2024stance}, KAI \citep{zhang2024knowledge}, EDDA with GPT (EDDA-GPT) and LLaMa (EDDA-LLaMa) \citep{ding2024edda}, Infuse \cite{yan2024collaborative}, and LKI-BART \citep{zhang2024llm}. The results of the following are taken directly from their original papers: LOT \cite{hu2024ladder}, RoBERTa-base-PV-P1 \citep{motyka2024target}, CNet-Ad \citep{zhang2024commonsense}, MCLDA \citep{wang2024meta}, and S-ESD \citep{ding2024distantly}.

\par \noindent {\textbf{EZ-STANCE:}} We implement the top three performing models from the original EZ dataset paper \cite{zhao2024ez} as baselines: BART-MNLI-e\textsubscript{p}, BART-MNLI-e, and BART-MNLI.

\par \noindent {\textbf{P-Stance:}} We include the methods that report results on the P-Stance dataset for in-target [Infuse \cite{yan2024collaborative} and LOT \cite{hu2024ladder}] and zero-shot [COLA \citep{lan2024stance} and KAI \citep{zhang2024knowledge}] stance task. %We also include BERTweet \cite{} and GPT-3.5+COT being used in \cite{hu2024ladder} and \citep{lan2024stance}

%demonstrate strong performance on VAST: , Infuse \cite{yan2024collaborative}, KAI \citep{zhang2024knowledge}, NPS4SD \citep{zhang2023twitter}, EDDA-GPT, EDDA-LLaMa \citep{ding2024edda}, and LKI-BART \citep{zhang2024llm}.

\par \noindent {\textbf{RFD:}} Due to the lack of available code, we report the baseline results as provided in the original RFD dataset paper \citep{saha2024stance}, including: MoLE \citep{DBLP:conf/emnlp/HardalovANA21} + $XSD_p$, MTDNN \citep{schiller2021stance} + $XSD_p$, and HSD \citep{sepulveda2021exploring} + $XSD_p$. However, we implement Infuse \cite{yan2024collaborative}, the top-performing method on the VAST dataset, as a baseline on RFD to enable a fair comparison with IRIS across five runs.

\section{Results and Analysis \label{sec_res}}
We begin by analyzing the performance of LLMs, followed by a comparison between baselines and our IRIS on VAST and EZ datasets for ZSSD. We then conduct ablation studies using our primary datasets, VAST and EZ. To further examine the interpretability of IRIS, we evaluate both implicit and explicit rationales using human judgments and automated metrics. Next, we assess IRIS’s generalizability on P-Stance and RFD datasets. Finally, we provide case studies to offer deeper insight into IRIS's predictive behavior.   
%, We then conduct ablation studies using our primary datasets, VAST with zero-shot targets and EZ STANCE with noun phrase targets to analyze the contribution of individual IRIS components, explore alternative ranking strategies, assess sensitivity to various parameters, and evaluate the impact of different embedding choices.
\begin{table}
\centering
\scalebox{0.70}{
\begin{tabular}{|l|c|c||c|c|}
\hline
{\textbf{Models}} &{\textbf{Technique}}  &{\textbf{VAST}} &{\textbf{EZ}}\\ \hline
% \multicolumn{11}{|c|}{\textbf{\textit{With Large Language Models (LLMs)}}} \\ \hline

\multirow{3}{*} \textbf{Mistral}
&\textbf{Zero-shot(ZS)} &67.09	&42.58\\ 

&\textbf{Few-shot(FS)} &69.76	&44.06\\
&\textbf{{\begin{tabular}[c]{@{}l@{}}Fine-tune\end{tabular}}}  &71.77	&53.4	
\\ \hline

\multirow{3}{*} \textbf{{\begin{tabular}[c]{@{}l@{}}\textbf{Llama 3.1}\end{tabular}}} 
&\textbf{Zero-shot(ZS)} &63.03	&55.13
 \\ 

&\textbf{Few-shot(FS)} &67.55	&58.54

\\
&\textbf{{\begin{tabular}[c]{@{}l@{}}Fine-tune\end{tabular}}}  &\textbf{72.81}	&\textbf{63.28}

\\ \hline

\end{tabular}
}
\caption{Results of LLMs on VAST (zero-shot targets) and EZ (noun-phrase targets) for ZSSD.}
\label{tab_llm}
%\vspace{-0.4cm}
\end{table}
\subsection{Performance of LLMs} \label{sec_perf_llm}
Table \ref{tab_llm} presents the results of LLMs on VAST (zero-shot targets) and EZ (noun-phrase targets) datasets. It can be seen from Table \ref{tab_llm} that LLMs perform better when fine-tuned with the training datasets than zero- and few-shot settings. Llama 3.1 outperforms Mistral, showcasing its better alignment with the instructions compared to Mistral for the SD task. However, compared to other state-of-the-art methods, LLMs are still lagging (Tables \ref{tab_llm}, \ref{tab_base_vast}, and \ref{tab_base_ez}). Furthermore, we manually examined the $100$ randomly selected correct and incorrect predictions of Llama from both datasets and prompted it to provide the best subsequence of words leading to the predicted stance (best implicit rationale), which resulted in a significantly lower F1 score of 0.617 compared to the ground-truth (refer Section \ref{app_rationale} for analysis of $100$ samples). However, we observed satisfactory results when we prompted Llama to generate all possible implicit rationales and linguistic-based explicit rationales (detailed in Section \ref{sec_llm_prompts}). This motivated us to take advantage of the Llama model for generating implicit rationales as a subsequence of words and linguistic arguments rather than using LLMs for the task of recognizing attitudes directly. %and provide inherent interpretability in the stance detection framework that can extract the most relevant rationale leading to the correct stance. 
\begin{table}
\centering

\scalebox{0.65}{
\begin{tabular}{|l|l|l|l|l|}
\hline
{\textbf{Models}} &\multicolumn{4}{c|}{\textbf{VAST (Zero-shot)}}\\ \hline

&\textbf{Pro} & \textbf{Con} &\textbf{Neu} &\textbf{All} \\ \hline

\textbf{COLA$^\dag$} 
&71.85	&70.77	&80.38	&74.33
 \\ 

 \textbf{{\begin{tabular}[c]{@{}l@{}}EDDA-GPT$^\dag$\end{tabular}}} &67.5	&67.3	&89.6	&74.8

 \\ 

\textbf{{\begin{tabular}[c]{@{}l@{}}EDDA-LLaMA$^\dag$\end{tabular}}} 
&67.4	&71.2	&90.1	&76.23 \\

\textbf{{\begin{tabular}[c]{@{}l@{}}GPT-TiDA\end{tabular}}}  &-	&-  &-	&76.1 \\

\textbf{{\begin{tabular}[c]{@{}l@{}}KAI$^\dag$\end{tabular}}} 
&63.58	&71.29	&87.66	&74.17 \\

\textbf{{\begin{tabular}[c]{@{}l@{}}LKI-BART$^\dag$\end{tabular}}}
&74.6	&72.8	&90.2	&79.2
\\

\textbf{Cnet-Ad} 
 &63.3 &65.3 &{{91.1}} &73.2
\\
\textbf{MCLDA} 
&{67.0} &{68.2} &90.1 &{75.1}
\\

\textbf{S-ESD} 
&{62.0} &{68.1} &88.6 &{72.9}
\\

\textbf{{\begin{tabular}[c]{@{}l@{}}LOT\end{tabular}}}  &-	&-  &-	&78.6 \\

\textbf{{\begin{tabular}[c]{@{}l@{}}RoBERTa-PV-P1\end{tabular}}}  &-	&-  &-	&77.6 \\

\textbf{{\begin{tabular}[c]{@{}l@{}}Infuse$^\dag$\end{tabular}}}  &74.7	&75.1	&\textbf{94.5}	&81.43
 
 \\ 
\hline
\multicolumn{5}{|c|}{\textbf{\textit{Our IRIS (\% of training data)}}} \\ \hline
  
\textbf{{\begin{tabular}[c]{@{}l@{}}IRIS (10\%)$^\dag$\end{tabular}}} &73.48 &75.63 &86.92	&78.68 \\

\textbf{{\begin{tabular}[c]{@{}l@{}}IRIS (30\%)$^\dag$\end{tabular}}} &\underline{77.19$^*$}	&\underline{79.28$^*$}	&90.19	&\underline{82.22} \\

\textbf{{\begin{tabular}[c]{@{}l@{}}IRIS (50\%)$^\dag$\end{tabular}}} &\textbf{81.15$^*$}	&\textbf{82.06$^*$}	&\underline{93.47}	&\textbf{85.56$^*$}
\\ \hline

\end{tabular}
}
\caption{Results of IRIS and baselines on VAST (zero-shot targets) for ZSSD. $\dag$ denotes baselines re-run over 5 rounds; results for the remaining are taken directly from their original papers (Section~\ref{sec_baseline}). Standard deviations are in Table~\ref{tab_base_vast_stdev}. $*$ indicates IRIS outperforms baselines$^\dag$  at p $<$ 0.05 (paired t-test). [\textbf{bold}: best, \underline{underline}: second best.]}
\label{tab_base_vast}
\end{table}

\begin{table*}[]
\centering
\scalebox{0.80}{
\begin{tabular}{|l|l|l|l||l|l|l||l|l|l|}
\hline
{\textbf{Train/Val}} &\multicolumn{3}{c|}{\textbf{Mixed targets (M)}} &\multicolumn{3}{c|}{\textbf{Noun-phrase targets (N)}} &\multicolumn{3}{c|}{\textbf{Claim targets (C)}}\\ \hline

{\textbf{Test}} &\textbf{M} &\textbf{N} &\textbf{C} &\textbf{M} &\textbf{N} &\textbf{C} &\textbf{M} &\textbf{N} &\textbf{C} \\ \hline

\textbf{{\begin{tabular}[c]{@{}l@{}}BART-MNLI\end{tabular}}}  &65.51/1.07	&31.14/0.42	&79.51/1.10		&64.29/1.19	&32.24/2.01	&\textbf{81.45/0.16}	&67.75/1.08	&31.05/2.14	&80.44/1.02

 \\

\textbf{{\begin{tabular}[c]{@{}l@{}}BART-MNLI-e\end{tabular}}}  &79.29/1.15	&66.24/0.75	&87.43/0.69		&45.4/0.35	&67.06/0.25	&30.05/2.31
&\underline{71.9/0.13}	&35.46/1.35	&\underline{88.72/0.50}
 \\
\textbf{{\begin{tabular}[c]{@{}l@{}}BART-MNLI-e$_p$\end{tabular}}}  &\underline{81.01/0.54}	&\underline{65.48/0.71}	&\underline{88.37/0.65}		&44.81/0.15	&67.49/1.21	&32.52/0.26		&- &- &-

 \\

\hline
\multicolumn{10}{|c|}{\textbf{\textit{Our IRIS (\% of training data)}}} \\ \hline
  
\textbf{{\begin{tabular}[c]{@{}l@{}}IRIS (10\%)\end{tabular}}} &75.37/1.49	&60.68/2.16	&82.03/2.24		&65.05/1.31	&67.53/1.11	&58.13/0.72		&66.28/0.56	&50.22/2.45	&79.42/2.25

\\

\textbf{{\begin{tabular}[c]{@{}l@{}}IRIS (30\%)\end{tabular}}} &80.01/1.26	&65.43/2.02	&86.39/1.57		&\underline{69.11/2.15$^*$}	&\underline{71.34/1.15$^*$}	&62.28/1.5		&69.91/1.67	&\underline{55.47/2.31$^*$}	&85.38/1.19

\\
\textbf{{\begin{tabular}[c]{@{}l@{}}IRIS (50\%)\end{tabular}}} &\textbf{82.18/0.75}	&\textbf{71.15/1.11$^*$}	&\textbf{90.58/2.09}		&\textbf{72.38/2.03$^*$}	&\textbf{74.32/1.02$^*$} &\underline{67.29/1.11}		&\textbf{75.50/1.05$^*$}	&\textbf{60.09/0.35$^*$}	&\textbf{89.46/1.63}

\\ \hline

\end{tabular}
}
\caption{Results (Average/Std. Dev) of IRIS and baselines on EZ for ZSSD. All methods are run over 5 rounds. $*$ denotes IRIS outperforms baselines at p $<$ 0.05 with paired t-test. [\textbf{bold}: best, \underline{underline}: second best.]}
\label{tab_base_ez}
\end{table*}

\begin{comment}
\begin{table}
\centering

\scalebox{0.80}{
\begin{tabular}{|l|l|l|l|l|}
\hline
{\textbf{Models}} &\multicolumn{4}{c|}{\textbf{EZ (noun-phrase targets)}}\\ \hline

&\textbf{Against} & \textbf{Favor} &\textbf{None} &\textbf{All} \\ \hline

\textbf{{\begin{tabular}[c]{@{}l@{}}BART-MNLI-e$_p$\end{tabular}}}  &74	&72.4	&59.7	&68.70
 \\
\textbf{{\begin{tabular}[c]{@{}l@{}}BART-MNLI-e$_p$\end{tabular}}}  &- &- &-	&67.5
 \\

 \textbf{{\begin{tabular}[c]{@{}l@{}}RoBERTa-MNLI$_p$\end{tabular}}}  &- &- &-	&66.2
 \\

\textbf{{\begin{tabular}[c]{@{}l@{}}RoBERTa-MNLI\end{tabular}}}  &- &- &-	&65.9
 \\
 \textbf{{\begin{tabular}[c]{@{}l@{}}RoBERTa\end{tabular}}}  &- &- &-	&65.6
 \\
\hline
\multicolumn{5}{|c|}{\textbf{\textit{Our IRIS (\% of training data)}}} \\ \hline
  
\textbf{{\begin{tabular}[c]{@{}l@{}}IRIS (10\%)\end{tabular}}} &76.2	&72.06	&54.34	&67.53
\\

\textbf{{\begin{tabular}[c]{@{}l@{}}IRIS (30\%)\end{tabular}}} &78.05	&74.67	&61.3	&71.34
\\
\textbf{{\begin{tabular}[c]{@{}l@{}}IRIS (50\%)\end{tabular}}} &\textbf{81.09}	&\textbf{77.61}	&\textbf{64.28}	&\textbf{74.32}

\\ \hline

\end{tabular}
}
\caption{Results of IRIS and baselines on EZ.}
\label{tab_base_ez}
\end{table}
\end{comment}
\begin{figure}
\centering
\includegraphics[width=0.60\linewidth]{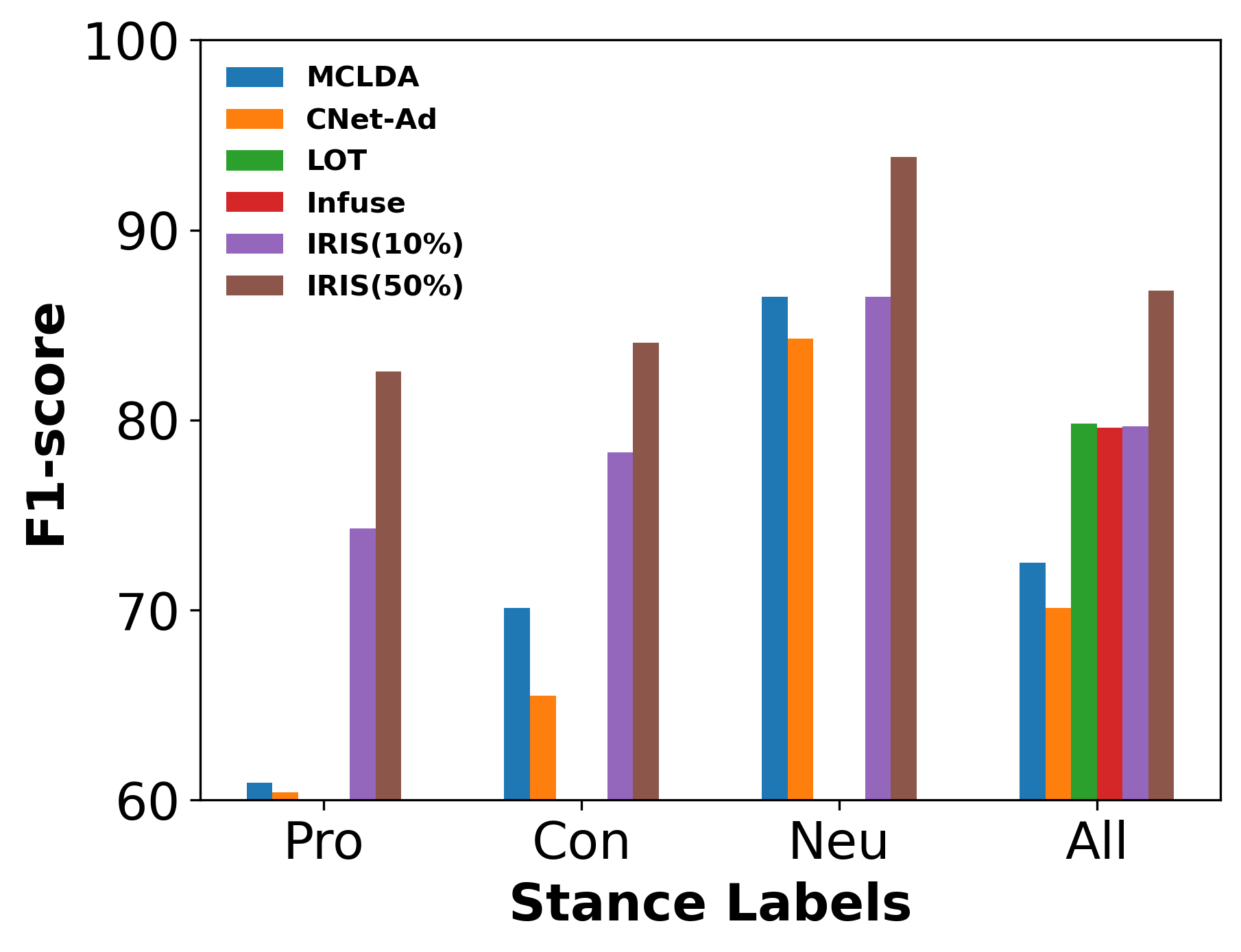}
     \caption{Results of IRIS with few-shot targets VAST}
  \label{fig_vast_few}
\end{figure}
\subsection{Comparison with Baselines (ZSSD)}
\subsubsection{VAST}
Table \ref{tab_base_vast} indicates that our IRIS approach trained on $50\%$ data outperforms other baselines with an overall averaged F1-score of $85.56$ over 5 rounds of execution, resulting in an average improvement of $12.62\%$ on the VAST for ZSSD. The standard deviation of IRIS and re-run baselines are reported in Table \ref{tab_base_vast_stdev}. We also report the results for different variants of our IRIS with $10\%$ and $30\%$ training data. Our IRIS ($10\%$) also performs better than most baselines on VAST data with the exception of LKI-BART and Infuse, which either used LLM to analyze the author’s implied emotions or infused target-related knowledge using the web. However, in contrast to these works, our IRIS prioritizes interpretability by automatically identifying relevant word subsequences and leveraging external documents to understand the contextual relationships. This helps in guiding the model toward accurate stances since we do not rely on ground truth for rationales. Thus, it surpasses LKI-BART and Infuse even with $30\%$ and $50\%$ training data. \textbf{\textit{Few-Shot Stance Detection}} Figure \ref{fig_vast_few} shows that our IRIS ($50\%$) significantly outperforms the other baseline methods with an overall F1 result of $86.81$ when tested with few-shot targets from VAST due to the efficient architecture guided by relevant rationale learning. It can be further seen that our IRIS trained on $10\%$ data with F1 score of $79.68$ outperforms the meta-contrastive learning-based MCDLA ($72.5$ F1) and common-sense based adversarial learning CNet-Ad ($70.1$ F1) and also results in comparable performance with Infuse ($79.60$ F1) and LOT ($79.80$ F1), thus highlighting the importance of our proposed ranking and selection stages that not only provide inherent interpretability, but lead the model to correct predictions, resulting in better performance even when trained with a smaller amount of data.

\subsubsection{EZ} Table~\ref{tab_base_ez} presents the average F1 scores and standard deviations of IRIS and all baselines over 5 runs. IRIS (50\%) consistently outperforms all baselines across nearly all train→test target combinations and ranks second-best in one case, demonstrating strong and robust performance (Table \ref{tab_base_ez}). Training on mixed targets and evaluating on specific ones (M→N and M→C) results in better performance with $71.15$ and $90.58$ F1 scores, compared to cross-target settings such as N→C and C→N with $67.29$ and $60.09$ F1, thus reflecting similar trends observed in baselines. This indicates that IRIS also benefits from mixed-target training data that includes both noun and claim targets. Further, results for M→N and M→C are also comparable to those for N→N and C→C (Table~\ref{tab_base_ez}), demonstrating IRIS's ability to generalize well under limited data. One exception is BART-MNLI, which outperforms IRIS on N→C, suggesting a need for more diverse training targets. This may be due to IRIS's reduced ability to transfer from noun-based training data to claim-based test inputs, possibly limited by incorrect rationales in relevance ranking stage. Nevertheless, IRIS achieves the best results on C→C ($89.46$ F1) and M→C ($90.58$ F1) when trained on claim and mixed targets and evaluated on claim targets. Furthermore, even with only $10\%$ or $30\%$ of training data, IRIS outperforms all baselines in combinations such as N→M, N→N, and C→N, and exceeds the performance of BART-MNLI on various combinations of M→M, M→N, M→C, N→M, N→N, and C→N, highlighting the effectiveness of combining relevance ranking and linguistic features for stance detection when trained
with a smaller amount of data.

\begin{table}
\centering
\scalebox{0.70}{
\begin{tabular}{|l|l||l|}
\hline
\textbf{Model} & \textbf{VAST} &\textbf{EZ} \\ \hline
\textbf{{\begin{tabular}[c]{@{}l@{}}Rational Gen. (RG) [Mistral]+\\Relevance Ranking (RR)\end{tabular}}} &73.02/2.49 &55.69/2.13 \\\hline 

\textbf{{\begin{tabular}[c]{@{}l@{}}RG [Llama] + RR\end{tabular}}} &79.43/1.86 &68.01/2.05 \\\hline 

\textbf{{\begin{tabular}[c]{@{}l@{}}RG+RR+Grouping\&Selection (GS)\end{tabular}}} &83.06/1.47 &71.52/1.65 \\ \hline

\textbf{{\begin{tabular}[c]{@{}l@{}}RG+RR+GS+Classification(C)\\ $[$Majority(Rel. Imp., Irr. Imp., Exp.)$]$ \end{tabular}}} &84.26/0.76 &72.29/1.31 \\ \hline

 \textbf{{\begin{tabular}[c]{@{}l@{}}RG+RR+GS+C\\
 $[$Majority(Rel. Imp., Exp.)$]$ 
 [IRIS(50\%)]\end{tabular}}} &85.56/0.33 &74.32/1.02 \\ \hline

\end{tabular}
}
\caption{Results (Avg./Std. Dev) for different IRIS components on VAST (zero-shot) and EZ (noun-phrase) for ZSSD.}
\label{tab_ablation_comp}
\end{table}

\subsection{Ablation Experiments}
We conduct ablation studies to analyze the contribution of individual IRIS components, explore alternative ranking strategies, assess sensitivity to various parameters, and evaluate the impact of different embedding choices. For ease of comparison, we focus on the primary datasets of VAST with zero-shot targets and EZ with noun phrase targets as part of our ablation experiments. We report average F1 metrics and standard deviations over $5$ runs.
\par \noindent \textbf{Significance of Different IRIS Components} Table \ref{tab_ablation_comp} presents the ablation study that justifies the importance of the different components of our IRIS approach. It can be seen that the rationales generated by Llama lead to better performance than those generated by Mistral LLM. From Table \ref{tab_ablation_comp}, the addition of the ``Grouping and Selection" component with the ``Relevance Ranking" leads to an improvement of the overall F1 result by $4.57\%$ and $5.16\%$ on VAST and EZ datasets respectively because of segregating the relevant and irrelevant rationales and extracting the diverse relevant rationale. It is also shown that concatenating the linguistic evaluation of posts together with implicit relevant and irrelevant rationales improves IRIS performance with $84.26$ (VAST) and $72.29$ (EZ) F1 scores. Moreover, stance classification based on the majority vote of explicit and implicit relevant rationales further guides IRIS to classify attitudes effectively (refer Table \ref{tab_ablation_comp}).
\begin{table}
\centering
\scalebox{0.65}{
\begin{tabular}{|l|l||l|}
\hline
\textbf{Model} & \textbf{VAST} &\textbf{EZ} \\ \hline
\textbf{{\begin{tabular}[c]{@{}l@{}}Relevance Ranking (RR) [FlagReranker]\end{tabular}}} &79.43/1.86 &68.01/2.13 \\\hline 

\textbf{{\begin{tabular}[c]{@{}l@{}}RR $\sim$ 
 instruction\end{tabular}}} &77.06/1.01 &66.35/1.27 \\ \hline

\textbf{{\begin{tabular}[c]{@{}l@{}}RR $\sim$ target \end{tabular}}} &76.25/0.93 &65.72/1.58 \\ \hline

%\textbf{{\begin{tabular}[c]{@{}l@{}}RR $\sim$ CARR (ours) \end{tabular}}} &75.08 &64.81 \\ \hline

\textbf{{\begin{tabular}[c]{@{}l@{}}RR $\sim$ (Flag)+LLM scores \end{tabular}}} &75.79/2.55 &66.2/2.36 \\ \hline

\textbf{{\begin{tabular}[c]{@{}l@{}}RR $\sim$ (Flag)+LLM as ranker \end{tabular}}} &74.91/2.37 &65.59/2.19 \\ \hline

\textbf{{\begin{tabular}[c]{@{}l@{}}RR$\sim$ (Flag)+flashrank\end{tabular}}} &75.42/0.84 &73.50/1.22 \\ \hline

\textbf{{\begin{tabular}[c]{@{}l@{}}RR$\sim$ (Flag)+rank-bm25\end{tabular}}} &68.15/1.04 &59.72/1.46 \\ \hline
 
\textbf{{\begin{tabular}[c]{@{}l@{}}RR$\sim$ (Flag)+cosine sim\end{tabular}}} &65.30/0.57 &57.24/0.32 \\ \hline

\end{tabular}
}
\caption{Results (Avg./Std. Dev) of various ranking algorithms in ranking stage on VAST and EZ for ZSSD. }
\label{tab_ablation_rel}

\end{table}

\par \noindent \textbf{Different Ranking Algorithms \label{sec_res_ranking}} Table \ref{tab_ablation_rel} represents the results when the relevance ranking (RR) stage is used exclusively for ZSSD without grouping and selection and explicit rationales. It is noted that our RR stage alone performs better than fine-tuned LLMs, proving the significance of our approach and justifying better performance than relying only on LLMs for ZSSD (Table \ref{tab_llm}). We find out that removing instruction and target from the query and document reduces task performance by $1.81\%$ and $3.68\%$ on average for both datasets. This suggests that the instruction helps the rankers understand the hidden requirement of aligning the query with the given target based on how the sentences in the documents address their targets. Moreover, the same sentences in datasets belong to different attitudes towards different targets, indicating the need to capture the target information in the query. We investigate LLM as a ranker providing relevance score of query w.r.t our 3 articulated documents and also LLM directly providing probabilistic scores of given query towards favor, against, and neutral stance (Prompt \ref{fig_prompt_im}). From Table \ref{tab_ablation_rel}, FlagReranker outperforms LLM by leveraging its powerful architecture based on well-established language models and document relevance along with enhanced retrieval capability, thus benefitting the stance detection pipeline. We also explore the use of flashrank\footnote{https://github.com/PrithivirajDamodaran/FlashRank} cross-encoder model fine-tuned on Amazon shopping query dataset (model\_name=`ce-esci-MiniLM-L12-v2'), rank-bm25 \cite{rank25} and cosine similarity (torch.nn.functional.cosine\_similarity) instead of FlagReranker that we used as a ranking algorithm, and find that the FlagReranker was more effective than others because our datasets contain similar statements for different targets leading to different attitudes, and these methods failed to capture the implicit and underlying relationships of the statements with the targets.
\par \noindent \textbf{Sensitivity Analysis} \label{sec_sens}
\textit{Figure \ref{fig_k1}} plots the overall F1 score for different values of $k$, i.e. for the selection of $k$ relevant and $k$ irrelevant rationales extracted using the ``diverse rationale selection fallback" component. It can be seen that the selection of k=3 for VAST and EZ achieves superior performance. Beyond this point, there is a noticeable decline in performance, as the average number of rationales retrieved per input is $6.25$ for VAST and $4.15$ for EZ, possibly leading to more irrelevant or empty subgroups during the grouping and selection stage of IRIS. \textit{Figure \ref{fig_beta1}} plots the overall F1 score for different values of $\beta$ used as reward/punish metric for Rationale Usefulness Reward Punish loss function ($L_{rp}$). The parameter $\beta$ controls the trade-off between rewarding useful rationales and punishing less useful or irrelevant ones. According to various experimental analyses, it is found that $\beta$ = $0.1$ performs best for both VAST and EZ and was therefore chosen as the hyperparameter for our model. This could be due to how $\beta$ balances the reward and punishment dynamics in training. A value of $\beta$=$0.1$ encourages IRIS to explore and learn useful patterns without being overly constrained by penalties. If $\beta$ is larger (e.g. $\beta$ = $0.2$), the model is rewarded more, but at the same time the penalty for selecting less useful rationales becomes stronger, which could discourage the model from exploring fewer specific areas of the data and potentially missing informative but noisy patterns. The observed result when $\beta$ = $0.1$ strikes a balance in scenarios with competing objectives (reward vs. punishment), ensuring robust learning of useful rationales.
\par \noindent \textbf{Different Embeddings} Appendix \ref{app_embed} and Table \ref{tab_ablation_emb}.

%Tables \ref{tab_base_pstance} and \ref{tab_base_rfd} show that our IRIS when trained on $50\%$ data outperforms the baselines on both P-Stance for in-target and zero-shot stance and on RFD for in-target stance task. IRIS on P-Stance improves the baselines with $11.29\%$ and $7.83\%$ average improvement in F1 scores for in-target and zero-shot respectively. Further, our zero-shot IRIS superseeds the in-target baselines with $85.77$ F1 that are trained and tested on similar targets, however,our ranking and grouping stage helps in selection of relevant rationales without ground-truth improves the performance of stance task even in zero-shot settings. Similary, IRIS outperforms all baselines on RFD (Table \ref{tab_base_rfd}), indicating that IRIS can handle longer length conxtual articles due to advantage of focusing on linguistic interpretation together with the most relevant subsequence of words within the article to identify correct stance of the article.

\begin{table}
\centering

\scalebox{0.55}{
\begin{tabular}{|l|l|l|l|l|}
\hline
{\textbf{Rationales}} &\textbf{Sufficiency}  &\textbf{Comprehensiveness}   &\textbf{Faithfulness} &\textbf{Plausibility}\\ \hline

\textbf{{\begin{tabular}[c]{@{}l@{}}Implicit\\(subsequence of words)\end{tabular}}}  &4.23	&3.55	&4.36	&3.49\\ 
 
\textbf{{\begin{tabular}[c]{@{}l@{}}Explicit\\(linguistic measures)\end{tabular}}}  
&4.35	&3.89	&4.25	&4.37 \\ \hline 

\end{tabular}
}
\caption{Human evaluation of rationales extracted by IRIS}
\label{tab_eval_human}
\end{table}

\subsection {Evaluation of Rationales \label{sec_eval_rat}} To evaluate the interpretability of IRIS, we conduct both automatic and human evaluations of the implicit and explicit rationales extracted by IRIS ($50\%$) for decoding stance. \textbf{Setup:} We selected 100 random test samples from VAST (zero-shot) and EZ (noun-phrase). These are the same samples previously used to assess rationale quality from LLM, which initially motivated our use of more effective ranking algorithms within IRIS (Section \ref{app_rationale}). For \textit{implicit rationales}, ground truth annotations were already created by annotators, framing this as a binary classification task: each token is labeled as 1 (part of rationale) or 0 (not part of rationale) (Section \ref{app_rationale}). We use the relevant implicit rationales extracted by IRIS after the grouping and selection stage for evaluation. For \textit{explicit linguistic rationales}, the same team of $3$ annotators performed the linguistic annotations. They were provided the same linguistic rationale-generation prompt as used for LLMs (Figure \ref{fig_prompt_ex}). Two doctoral researchers annotated 50 samples each, while a third postdoctoral researcher reviewed all annotations, requesting revisions where necessary to ensure high-quality labels. \\
\textbf{Automatic Evaluation:} \textit{Implicit Rationales:} We evaluate IRIS extracted relevant implicit rationales against the ground truth using F1 as rationale identification is considered a binary classification task. IRIS achieves an F1 of 0.859, significantly outperforming Llama’s score of 0.617 (Section \ref{app_rationale}), demonstrating the efficacy of IRIS in identifying the most relevant subsequences for interpretability. \textit{Explicit linguistic Rationales:} We use BLEU \citep{DBLP:conf/acl/PapineniRWZ02} and BERTScore \citep{DBLP:conf/iclr/ZhangKWWA20} to evaluate the quality of explicit rationales in IRIS. We observe BLEU 1, BLEU 2, BLEU 3, and BLEU 4 scores of 21.1, 18.54, 11.08, and 5.71, respectively, reflecting n-gram precision from unigrams to four-grams, resulting in an average BLEU score of 14.10, while a BERTScore of 84.01 indicates strong semantic alignment between IRIS-generated and ground-truth rationales. Please note that we simply use LLM-generated linguistic measures as explicit rationales in IRIS, reviewed by annotators (Section \ref{sec_llm_prompts}), showing that while LLMs may struggle with classification, they are effective in generating high-quality textual explanations. \\
\textbf{Human Evaluation} A separate team of annotators consisting of $3$ master students with a computational social science background rated IRIS-generated implicit and explicit rationales on a 5-point Likert scale across four criteria: \textit{Sufficiency} (how well it justifies the prediction), \textit{Comprehensiveness} (coverage of relevant aspects), \textit{Faithfulness} (alignment with the model’s true reasoning), and \textit{Plausibility} (how convincing it is to a human), where 1 indicates poor and 5 indicates strong alignment. As shown in Table~\ref{tab_eval_human}, both rationale types scored similarly on sufficiency and comprehensiveness. Explicit (linguistic) rationales were more plausible due to their detailed explanations, while implicit rationales were rated higher in faithfulness, better reflecting the model’s reasoning (Table~\ref{tab_eval_human}). This highlights the complementary interpretability of both rationale types in IRIS.
\begin{table}
\centering

\scalebox{0.60}{
\begin{tabular}{|l|l|l|l|}
\hline
{\textbf{Models}} &\textbf{Trump}  &\textbf{Biden}  &\textbf{Sanders}\\ \hline

\multicolumn{4}{|c|}{\textbf{\textit{In-target Stance Detection}}} \\ \hline

\textbf{{\begin{tabular}[c]{@{}l@{}}LOT\end{tabular}}}  &86.1	&83.7	&80.5
 \\ 
 
\textbf{Infuse$^\dag$} 
&86.2/0.39	&84.5/0.95	&80.53/0.86	
 \\ 

%\textbf{{\begin{tabular}[c]{@{}l@{}}Llama 3.1 (fine-tune)\end{tabular}}}  &86.1	&83.7	&80.5 \\ 

\textbf{{\begin{tabular}[c]{@{}l@{}}IRIS (50\%)\end{tabular}}} &\textbf{90.49/2.51$^*$}	&\textbf{91.19/2.06$^*$}	&\textbf{89.61/2.19$^*$}	
\\  \hline

\multicolumn{4}{|c|}{\textbf{\textit{Zero-Shot Stance Detection (ZSSD)}}} \\ \hline

\textbf{{\begin{tabular}[c]{@{}l@{}}COLA$^\dag$\end{tabular}}}  &87.1/0.57	&84.6/0.35	&80.4/0.61 \\

\textbf{{\begin{tabular}[c]{@{}l@{}}KAI$^\dag$\end{tabular}}}  &72.45/2.05	&85.16/1.51	&78.69/1.73
 \\ 
%\textbf{{\begin{tabular}[c]{@{}l@{}}Llama 3.1 (zero-shot)\end{tabular}}}  &75.66/1.79	&79.08/2.15	&75.95/2.66 \\

\textbf{{\begin{tabular}[c]{@{}l@{}}IRIS (50\%)\end{tabular}}} &\textbf{87.98/1.15}	&\textbf{89.05/1.08$^*$}	&\textbf{85.77/0.82$^*$}	
\\ \hline

\end{tabular}
}
\caption{Results (Average/Std. Dev) of IRIS (50\%) and baselines on P-Stance for in-target and zero-shot stance. $\dag$ indicates baselines are re-run over 5 rounds; results for the remaining are taken directly from their original papers (Section \ref{sec_baseline}). $*$ denotes IRIS outperforms baselines$^\dag$ at p $<$ 0.05.}
\label{tab_base_pstance}
\end{table}

\begin{table}
\centering
\scalebox{0.60}{
\begin{tabular}{|l|l|}
\hline
\textbf{Model} & \textbf{Macro F1} \\ \hline
\textbf{Mole+XSD$_p$} &0.69 \\  

\textbf{MTDNN+XSD$_p$} &0.52 \\ 

\textbf{HSD+XSD$_p$} &0.40 \\ 

\textbf{Infuse$^\dag$} &0.67/2.02 \\ 

\textbf{IRIS (50\%)} &0.73/1.69$^*$ \\ \hline 

\end{tabular}
}
\caption{Results (Average/Std. Dev) of IRIS (50\%) and baselines on RFD for in-target stance. $\dag$ indicates baselines are re-run over 5 rounds; results for the remaining are taken directly from their original papers (Section \ref{sec_baseline}). $*$ denotes IRIS outperforms baselines$^\dag$ at p $<$ 0.05 (paired t-test).}
\label{tab_base_rfd}
\end{table}

\subsection{Generalizability Analysis}
Tables~\ref{tab_base_pstance} and~\ref{tab_base_rfd} demonstrate that IRIS, when trained on $50\%$ of the data, outperforms all baselines on both P-Stance (in-target and zero-shot) and RFD (in-target), respectively. On P-Stance, IRIS improves F1 scores by $11.29\%$ and $7.83\%$ in in-target and zero-shot stance tasks.
%On P-Stance, IRIS achieves an average improvement of $11.29\%$ and $7.83\%$ in F1 scores the in-target and zero-shot settings.
Notably, the zero-shot version of our IRIS surpasses even in-target baselines across all domains of P-Stance, highlighting the strength of our relevance ranking and rationale grouping stages, which enable effective rationale selection even without access to ground-truth labels. Similarly, on RFD (Table~\ref{tab_base_rfd}), IRIS outperforms all baselines, demonstrating its ability to manage longer contextual articles by leveraging linguistic cues and identifying the most relevant subsequences to determine the correct stance.

\subsection{Qualitative Analysis}
To better understand IRIS’s predictive capabilities for stance detection, we examine case studies in Table \ref{tab_qual_exam}. Examples 1–3 demonstrate correct predictions across different datasets, supported by relevant implicit rationales (subsequences of text) and explicit rationales (communicative text features). Example 3 from Table \ref{tab_qual_exam} shows that IRIS correctly identifies the implicit rationale, in contrast to the $XSD_p$ approach proposed in \citep{saha2024stance}, which misidentifies the relevant rationale for the same instance. Due to space constraints, we could not include Figure 7 that illustrates this error in the $XSD_p$ method (refer \citep{saha2024stance}). This example further highlights the strength of IRIS in handling longer and more complex texts, such as those in RFD, where multiple statements may imply different stances. IRIS effectively extracts the top k most relevant implicit rationales during the grouping and selection stage to guide accurate classification—where k is 7 for RFD due to its longer text length, compared to 3 for other datasets. Examples 4 and 5 illustrate error scenarios. In Example 4, IRIS focuses on positive sentiment toward “Smolensk” while referencing negative actions by others, resulting in a “favor” prediction. In Example 5, although the predicted stance is correct, the implicit rationales include redundancy and miss the most relevant justification when compared with human annotations. However, explicit rationales still provide sufficient reasoning, reinforcing the value of using both types of rationales in IRIS for accurate and interpretable stance prediction. To further understand the benefits of our prompts and IRIS approach, we looked into step-by-step processing for one of the sample predictions from VAST (Example 1 of Table \ref{tab_qual_exam}) by LLMs only and our IRIS model, detailed in Appendix \ref{app_qual}.

\section{Conclusion}
%\textcolor{red}{Make more exciting: explicitly highlight the practical implications or potential real-world applications of the framework. For example, you could add a sentence like: These advancements hold promise for real-world applications in areas such as online content moderation, sentiment analysis, and social media monitoring.}
In this work, we present a novel zero-shot stance detection system that offers an interpretable understanding of the attitude of the input by focusing on implicit sequence-based and explicit linguistic-based rationales. Our approach understands the context of the rationales toward the different stances while ensuring that the relevant rationales are captured to guide the model in correcting stance predictions. Extensive experiments on various benchmark datasets using ${50\%}$, ${30\%}$, and even ${10\%}$ training data validate the significance of the different components of our approach. By making our framework implicitly and explicitly interpretable, we ensure that the model is usable in complex environments like online content moderation, and social media monitoring, and help identify biased rationales in sensitive areas such as politics or climate change to ensure the ethical use of AI. In the future, we plan to explore these advances and focus on the interpretability of low-resource stance detection.

\section{Limitations}
Despite the considerable success we have achieved with our study, we must acknowledge the limitations of our work for future refinement. We have used datasets that are available in English. We may need to explore other embeddings and pre-trained rankers if used for other languages.
%other embeddings such as "bge-multilingual-gemma2\footnote{https://huggingface.co/BAAI/bge-multilingual-gemma2}" and others that have been shown to be effective in multilingual classification tasks. We also need to search for a multilingual pre-trained ranker algorithm. 
However, the basis of our approach, namely using relevance ranking and filtering out relevant and irrelevant diverse reasoning might still be suitable for multilingual or low-resource stance detection after we have optimized some embeddings. %We consider this as part of further work and can therefore consider it as a limitation for the current method. 
As we focus on target-based stance, we can expand our scope to target-independent and domain-based stance detection. Further enhancement of the quality of our external knowledge base with higher quality assertions of stance can improve the task. %We also plan to integrate a target detection module in the future that can automatically identify all implicit or explicit targets in a post and then determine the attitude towards these targets instead of relying on target information. These steps can help improve the performance of the relevance ranking stage and steer the model in the right direction when there is no ground-truth for rationales. 
It might be possible that incorrect predictions can influence public opinion or decision-making processes in critical social issues, for instance, our approach can be used by political parties during their election campaigns to regularly monitor how public opinion on social issues such as COVID or climate change has shifted either for or against them in order to gain an advantage for their own interests. Therefore, in our work, we aim to achieve interpretability. However, further improving the performance of the relevance ranking stage could steer the model in the right direction. Consequently, future work should consider these shortcomings and work towards the refinement process.

\bibliography{aaai25}

\begin{thebibliography}{38}
\providecommand{\natexlab}[1]{#1}

\bibitem[{Allaway and McKeown(2020)}]{DBLP:conf/emnlp/AllawayM20}
Allaway, E.; and McKeown, K.~R. 2020.
\newblock Zero-Shot Stance Detection: {A} Dataset and Model using Generalized Topic Representations.
\newblock In \emph{Proceedings of the 2020 Conference on Empirical Methods in Natural Language Processing, {EMNLP}}. Association for Computational Linguistics.

\bibitem[{Bergstra, Yamins, and Cox(2013)}]{bergstra2013making}
Bergstra, J.; Yamins, D.; and Cox, D. 2013.
\newblock Making a science of model search: Hyperparameter optimization in hundreds of dimensions for vision architectures.
\newblock In \emph{International conference on machine learning}, 115--123. PMLR.

\bibitem[{Boyd et~al.(2022)Boyd, Ashokkumar, Seraj, and Pennebaker}]{boyd2022development}
Boyd, R.~L.; Ashokkumar, A.; Seraj, S.; and Pennebaker, J.~W. 2022.
\newblock The development and psychometric properties of LIWC-22.
\newblock \emph{Austin, TX: University of Texas at Austin}, 10.

\bibitem[{Brown(2020)}]{rank25}
Brown, D. 2020.
\newblock Rank-BM25: A Collection of BM25 Algorithms in Python.
\newblock \emph{https://pypi.org/project/rank-bm25/}.

\bibitem[{Chen et~al.(2024)Chen, Zhou, Hua, Loh, Chen, Li, Zhu, and Liang}]{chen2024fintextqa}
Chen, J.; Zhou, P.; Hua, Y.; Loh, Y.; Chen, K.; Li, Z.; Zhu, B.; and Liang, J. 2024.
\newblock FinTextQA: A Dataset for Long-form Financial Question Answering.
\newblock \emph{arXiv preprint arXiv:2405.09980}.

\bibitem[{Devlin et~al.(2019)Devlin, Chang, Lee, and Toutanova}]{devlin2019bert}
Devlin, J.; Chang, M.-W.; Lee, K.; and Toutanova, K. 2019.
\newblock Bert: Pre-training of deep bidirectional transformers for language understanding.
\newblock In \emph{Proceedings of the North American chapter of the association for computational linguistics: human language technologies}.

\bibitem[{DeYoung et~al.(2020)DeYoung, Jain, Rajani, Lehman, Xiong, Socher, and Wallace}]{deyoung2019eraser}
DeYoung, J.; Jain, S.; Rajani, N.~F.; Lehman, E.; Xiong, C.; Socher, R.; and Wallace, B.~C. 2020.
\newblock {ERASER:} {A} Benchmark to Evaluate Rationalized {NLP} Models.
\newblock In \emph{Proceedings of the 58th Annual Meeting of the Association for Computational Linguistics, {ACL}}.

\bibitem[{Ding et~al.(2024{\natexlab{a}})Ding, Dai, Peng, Peng, Zhang, and Huang}]{ding2024distantly}
Ding, D.; Dai, G.; Peng, C.; Peng, X.; Zhang, B.; and Huang, H. 2024{\natexlab{a}}.
\newblock Distantly Supervised Explainable Stance Detection via Chain-of-Thought Supervision.
\newblock \emph{Mathematics}, 12(7).

\bibitem[{Ding et~al.(2024{\natexlab{b}})Ding, Dong, Huang, Xu, Huang, Liu, Jing, and Zhang}]{ding2024edda}
Ding, D.; Dong, L.; Huang, Z.; Xu, G.; Huang, X.; Liu, B.; Jing, L.; and Zhang, B. 2024{\natexlab{b}}.
\newblock EDDA: A Encoder-Decoder Data Augmentation Framework for Zero-Shot Stance Detection.
\newblock \emph{arXiv preprint arXiv:2403.15715}.

\bibitem[{Guo, Jiang, and Liao(2024)}]{guo2024improving}
Guo, M.; Jiang, X.; and Liao, Y. 2024.
\newblock Improving Zero-Shot Stance Detection by Infusing Knowledge from Large Language Models.
\newblock In \emph{International Conference on Intelligent Computing}, 121--132. Springer.

\bibitem[{Hardalov et~al.(2021)Hardalov, Arora, Nakov, and Augenstein}]{DBLP:conf/emnlp/HardalovANA21}
Hardalov, M.; Arora, A.; Nakov, P.; and Augenstein, I. 2021.
\newblock Cross-Domain Label-Adaptive Stance Detection.
\newblock In Moens, M.; Huang, X.; Specia, L.; and Yih, S.~W., eds., \emph{Proceedings of the 2021 Conference on Empirical Methods in Natural Language Processing, {EMNLP}}. ACL.

\bibitem[{Hu et~al.(2024)Hu, Yan, Chong, Yap, Guan, Zhou, and Tsang}]{hu2024ladder}
Hu, K.; Yan, M.; Chong, W.~H.; Yap, Y.~K.; Guan, C.; Zhou, J.~T.; and Tsang, I.~W. 2024.
\newblock Ladder-of-thought: Using knowledge as steps to elevate stance detection.
\newblock In \emph{2024 International Joint Conference on Neural Networks (IJCNN)}.

\bibitem[{Kumar(2024)}]{kumar2024overriding}
Kumar, S. 2024.
\newblock Overriding Safety protections of Open-source Models.
\newblock \emph{arXiv preprint arXiv:2409.19476}.

\bibitem[{Lan et~al.(2024)Lan, Gao, Jin, and Li}]{lan2024stance}
Lan, X.; Gao, C.; Jin, D.; and Li, Y. 2024.
\newblock Stance detection with collaborative role-infused llm-based agents.
\newblock In \emph{Proceedings of the International AAAI Conference on Web and Social Media}, volume~18, 891--903.

\bibitem[{Lee et~al.(2024)Lee, Roy, Xu, Raiman, Shoeybi, Catanzaro, and Ping}]{lee2024nv}
Lee, C.; Roy, R.; Xu, M.; Raiman, J.; Shoeybi, M.; Catanzaro, B.; and Ping, W. 2024.
\newblock NV-Embed: Improved Techniques for Training LLMs as Generalist Embedding Models.
\newblock \emph{arXiv preprint arXiv:2405.17428}.

\bibitem[{Lewis et~al.(2019)Lewis, Liu, Goyal, Ghazvininejad, Mohamed, Levy, Stoyanov, and Zettlemoyer}]{lewis2019bart}
Lewis, M.; Liu, Y.; Goyal, N.; Ghazvininejad, M.; Mohamed, A.; Levy, O.; Stoyanov, V.; and Zettlemoyer, L. 2019.
\newblock Bart: Denoising sequence-to-sequence pre-training for natural language generation, translation, and comprehension.
\newblock \emph{arXiv preprint arXiv:1910.13461}.

\bibitem[{Li et~al.(2021)Li, Sosea, Sawant, Nair, Inkpen, and Caragea}]{li2021p}
Li, Y.; Sosea, T.; Sawant, A.; Nair, A.~J.; Inkpen, D.; and Caragea, C. 2021.
\newblock P-stance: A large dataset for stance detection in political domain.
\newblock In \emph{Findings of the Association for Computational Linguistics: ACL-IJCNLP 2021}.

\bibitem[{Ma et~al.(2024)Ma, Wang, Xing, Zhao, and Zhang}]{ma2024chain}
Ma, J.; Wang, C.; Xing, H.; Zhao, D.; and Zhang, Y. 2024.
\newblock Chain of Stance: Stance Detection with Large Language Models.
\newblock In \emph{Natural Language Processing and Chinese Computing - 13th Conference, {NLPCC}}. Springer.

\bibitem[{Motyka and Piasecki(2024)}]{motyka2024target}
Motyka, D.; and Piasecki, M. 2024.
\newblock Target-Phrase Zero-Shot Stance Detection: Where Do We Stand?
\newblock In \emph{International Conference on Computational Science}. Springer.

\bibitem[{Papineni et~al.(2002)Papineni, Roukos, Ward, and Zhu}]{DBLP:conf/acl/PapineniRWZ02}
Papineni, K.; Roukos, S.; Ward, T.; and Zhu, W. 2002.
\newblock Bleu: a Method for Automatic Evaluation of Machine Translation.
\newblock In \emph{Proceedings of the 40th Annual Meeting of the Association for Computational Linguistics}.

\bibitem[{Reimers and Gurevych(2019)}]{reimers2019sentence}
Reimers, N.; and Gurevych, I. 2019.
\newblock Sentence-bert: Sentence embeddings using siamese bert-networks.
\newblock \emph{arXiv preprint arXiv:1908.10084}.

\bibitem[{Saha, Lakshmanan, and Ng(2024)}]{saha2024stance}
Saha, R.~R.; Lakshmanan, L.~V.; and Ng, R.~T. 2024.
\newblock Stance Detection with Explanations.
\newblock \emph{Computational Linguistics}.

\bibitem[{Schiller, Daxenberger, and Gurevych(2021)}]{schiller2021stance}
Schiller, B.; Daxenberger, J.; and Gurevych, I. 2021.
\newblock Stance detection benchmark: How robust is your stance detection?
\newblock \emph{KI-K{\"u}nstliche Intelligenz}, 1--13.

\bibitem[{Sep{\'u}lveda-Torres et~al.(2021)Sep{\'u}lveda-Torres, Vicente, Saquete, Lloret, and Palomar}]{sepulveda2021exploring}
Sep{\'u}lveda-Torres, R.; Vicente, M.; Saquete, E.; Lloret, E.; and Palomar, M. 2021.
\newblock Exploring summarization to enhance headline stance detection.
\newblock In \emph{International Conference on Applications of Natural Language to Information Systems}, 243--254. Springer.

\bibitem[{Singh et~al.(2024)Singh, Inala, Galley, Caruana, and Gao}]{singh2024rethinking}
Singh, C.; Inala, J.~P.; Galley, M.; Caruana, R.; and Gao, J. 2024.
\newblock Rethinking interpretability in the era of large language models.
\newblock \emph{arXiv preprint arXiv:2402.01761}.

\bibitem[{Taranukhin, Shwartz, and Milios(2024)}]{taranukhin2024stance}
Taranukhin, M.; Shwartz, V.; and Milios, E.~E. 2024.
\newblock Stance Reasoner: Zero-Shot Stance Detection on Social Media with Explicit Reasoning.
\newblock In \emph{Proceedings of the 2024 Joint International Conference on Computational Linguistics, Language Resources and Evaluation, {LREC/COLING}}.

\bibitem[{Upadhyaya, Fisichella, and Nejdl(2023)}]{upadhyaya2023toxicity}
Upadhyaya, A.; Fisichella, M.; and Nejdl, W. 2023.
\newblock Toxicity, morality, and speech act guided stance detection.
\newblock In \emph{Findings of the Association for Computational Linguistics: EMNLP 2023}, 4464--4478.

\bibitem[{Wang, Zhang, and Wang(2024)}]{wang2024meta}
Wang, C.; Zhang, Y.; and Wang, S. 2024.
\newblock A meta-contrastive learning with data augmentation framework for zero-shot stance detection.
\newblock \emph{Expert Systems with Applications}, 123956.

\bibitem[{Xiao et~al.(2023)Xiao, Liu, Zhang, Muennighoff, Lian, and Nie}]{xiao2023c}
Xiao, S.; Liu, Z.; Zhang, P.; Muennighoff, N.; Lian, D.; and Nie, J.-Y. 2023.
\newblock C-pack: Packaged resources to advance general chinese embedding.
\newblock \emph{arXiv preprint arXiv:2309.07597}.

\bibitem[{Yan, Joey, and Ivor(2024)}]{yan2024collaborative}
Yan, M.; Joey, T.~Z.; and Ivor, W.~T. 2024.
\newblock Collaborative knowledge infusion for low-resource stance detection.
\newblock \emph{Big Data Mining and Analytics}, 7(3): 682--698.

\bibitem[{Yang(2024)}]{yang2024diagnosing}
Yang, X. 2024.
\newblock Diagnosing Hate Speech Classification: Where Do Humans and Machines Disagree, and Why?
\newblock \emph{arXiv preprint arXiv:2410.10153}.

\bibitem[{Yao, Yang, and Wei(2024)}]{yao2024enhancing}
Yao, Z.; Yang, W.; and Wei, F. 2024.
\newblock Enhancing Zero-Shot Stance Detection with Contrastive and Prompt Learning.
\newblock \emph{Entropy}, 26(4): 325.

\bibitem[{Zhang et~al.(2024{\natexlab{a}})Zhang, Ding, Huang, Li, Li, Zhang, and Huang}]{zhang2024knowledge}
Zhang, B.; Ding, D.; Huang, Z.; Li, A.; Li, Y.; Zhang, B.; and Huang, H. 2024{\natexlab{a}}.
\newblock Knowledge-Augmented Interpretable Network for Zero-Shot Stance Detection on Social Media.
\newblock \emph{IEEE Transactions on Computational Social Systems}.

\bibitem[{Zhang et~al.(2024{\natexlab{b}})Zhang, Li, Zhu, and Li}]{zhang2024commonsense}
Zhang, H.; Li, Y.; Zhu, T.; and Li, C. 2024{\natexlab{b}}.
\newblock Commonsense-based adversarial learning framework for zero-shot stance detection.
\newblock \emph{Neurocomputing}, 563: 126943.

\bibitem[{Zhang et~al.(2020)Zhang, Kishore, Wu, Weinberger, and Artzi}]{DBLP:conf/iclr/ZhangKWWA20}
Zhang, T.; Kishore, V.; Wu, F.; Weinberger, K.~Q.; and Artzi, Y. 2020.
\newblock BERTScore: Evaluating Text Generation with {BERT}.
\newblock In \emph{8th International Conference on Learning Representations, {ICLR}}.

\bibitem[{Zhang et~al.(2024{\natexlab{c}})Zhang, Li, Zhang, and Xu}]{zhang2024llm}
Zhang, Z.; Li, Y.; Zhang, J.; and Xu, H. 2024{\natexlab{c}}.
\newblock LLM-Driven Knowledge Injection Advances Zero-Shot and Cross-Target Stance Detection.
\newblock In \emph{Proceedings of the 2024 Conference of the North American Chapter of the Association for Computational Linguistics: Human Language Technologies (Volume 2: Short Papers)}, 371--378.

\bibitem[{Zhao and Caragea(2024)}]{zhao2024ez}
Zhao, C.; and Caragea, C. 2024.
\newblock EZ-STANCE: A Large Dataset for English Zero-Shot Stance Detection.
\newblock In \emph{Proceedings of the 62nd Annual Meeting of the Association for Computational Linguistics (Volume 1: Long Papers)}.

\bibitem[{Zhao et~al.(2024)Zhao, Ma, Pang, Guo, Zhao, and Miao}]{zhao2024zero}
Zhao, X.; Ma, G.; Pang, S.; Guo, Y.; Zhao, J.; and Miao, J. 2024.
\newblock Zero-shot stance detection based on multi-expert collaboration.
\newblock \emph{Scientific Reports}, 14(1): 18092.

\end{thebibliography}

\section*{Ethics Statement}
All the datasets that we utilize for this research are open-access datasets. The  VAST and EZ-Stance datasets provide full-text data directly.
\par We acknowledge the potential risks associated with the misuse of our technology, as is the case with many innovations. Specifically, there is a possibility that it could be leveraged for unethical purposes, such as suppressing critics or identifying and targeting dissenting voices on social media by certain entities. We strongly encourage users of our technology to adhere to principles of ethical and responsible use.

\appendix
\begin{algorithm}
\caption{Diverse Rationale Selection Fallback}
\label{alg_div1}
\begin{algorithmic}[1] % The [1] ensures line numbers start from 1
\Statex \textbf{Dynamic Target Distribution:} 
\State  Let $N_f, N_a, N_n$ be the number of available rationales in the favor, against, and neutral groups, respectively.
\State The total number of rationales in the group (relevant or irrelevant) is: $N_{t} = N_f+N_a+N_n$
\State The target distribution $V$ is computed as
\Statex  \hspace{1em}  
$V$ =$(\frac{N_{f}}{N_{t}},\frac{N_{a}}{N_{t}},\frac{N_{n}}{N_{t}})$

\Statex \textbf{KL-Divergence Based Selection}
\Statex \textbf{Inputs:} $G$: set of relevant/irrelevant rationales; $k$: the number of relevant and irrelevant rationales to be selected; $V$: Target distribution for diversity across subgroups.
\Statex \textbf{Outputs:} $G_k$: A set of k selected rationales with diverse coverage over the subgroups.

\State Set Initial Distribution: Since no rationales are initially selected, the current choice distribution is initialized with $U_{current}= (\epsilon,\epsilon,\epsilon)$, where $\epsilon=10^{-6}$ to avoid division by zero in the KL-Divergence calculation.
\State Initialize selected rationales set: $G_k=\{\}$
\While {$|G_k| < k$}
    \For {candidate $r \in \mathcal R $}
    \State Add the candidate rationale $r$ to the selected set temporarily: $U_{new}= U_{current}$ +  update counts of favor, against, neutral subgroups.
    \State Calculate KL-divergence between the new distribution and the target distribution: $D_k (U_{new}||Q)= \sum_{j \in \{f,a,n\}} U_{new} (j) log(\frac{U_{new}(j)}{V(j)})$
    \State Select the rationale $r^*$ that that minimizes the KL-divergence: $r^*= \operatorname*{argmin}_r D_k (U_{new}||V)$
    \State Update the selected rationales: $G_k=G_k  \cup {r^*}$
    \State Update the distribution $U_{current}$ based on the selected rationale's subgroup (favor, against, neutral)
    \State  If a subgroup has no rationales left, skip it, but continue selecting from other subgroups.
    \EndFor
    \State \textbf{Fallback:} If it is not possible to maintain diversity (e.g., one of the subgroups is empty), continue selecting from the available groups to reach the total count $k$.
\EndWhile 
\State \textbf{return} $G_k$.
\Statex Process is repeated for both relevant and irrelevant sets of rationales. 
\State Here the output $G_k$ will be denoted by $Re_k$ in case of k relevant $Re_1, Re_2...Re_k$ set of implicit rationales and $Ie_k$ for the case of k irrelevant $Ie_1, Ie_2...Ie_k$ implicit rationales.
\end{algorithmic}
\end{algorithm}

\begin{table}
\centering

\scalebox{0.80}{
\begin{tabular}{|l|l|l|l|l|}
\hline
{\textbf{Models}} &\multicolumn{4}{c|}{\textbf{VAST (Zero-shot)}}\\ \hline

&\textbf{Pro} & \textbf{Con} &\textbf{Neu} &\textbf{All} \\ \hline

\textbf{COLA} 
&0.42	&0.78	&1.01	&0.45
 \\ 

\textbf{{\begin{tabular}[c]{@{}l@{}}EDDA-GPT\end{tabular}}} &0.29	&0.11	&0.64	&0.24

 \\ 

\textbf{{\begin{tabular}[c]{@{}l@{}}EDDA-LLaMA\end{tabular}}} 
&0.52	&0.33	&0.29	&0.39 \\

\textbf{{\begin{tabular}[c]{@{}l@{}}KAI\end{tabular}}} 
&1.44	&1.61	&2.13	&1.01 \\

\textbf{{\begin{tabular}[c]{@{}l@{}}LKI-BART\end{tabular}}}
&1.17	&0.61	&0.37	&0.79
\\

\textbf{{\begin{tabular}[c]{@{}l@{}}Infuse\end{tabular}}}  &0.42	&1.35	&0.54	&0.54
 
 \\ 
\hline
\multicolumn{5}{|c|}{\textbf{\textit{Our IRIS (\% of training data)}}} \\ \hline
  
\textbf{{\begin{tabular}[c]{@{}l@{}}IRIS (10\%)\end{tabular}}} &3.10	&3.09	&2.23	&1.68 \\

\textbf{{\begin{tabular}[c]{@{}l@{}}IRIS (30\%)\end{tabular}}} &2.01	&1.55	&1.69	&1.02 \\

\textbf{{\begin{tabular}[c]{@{}l@{}}IRIS (50\%)\end{tabular}}} &0.77	&0.36	&0.50	&0.33
\\ \hline

\end{tabular}
}
\caption{Standard deviation of IRIS and baselines run over 5 rounds on VAST with zero-shot targets.}
\label{tab_base_vast_stdev}
\end{table}

\begin{table}
\centering
\scalebox{0.60}{
\begin{tabular}{|l|l|}
\hline
\textbf{Model} &\textbf{Training Time} \\ \hline
\textbf{Infuse} &32.66 (1960) \\  

\textbf{KAI} &26.38 (1583) \\ 
\textbf{EDDA} &13.25 (795) \\ 

\textbf{LKI-BART } &23.13 (1388) \\ 

%\textbf{Rational Generation (RG)} &{\begin{tabular}[c]{@{}l@{}}Approx 1-2 hours for\\all datasets(pre-processing) \end{tabular}} \\

\textbf{Rational Generation (RG)+ Relevance Ranking(RR)} &18.41 (1105) \\

\textbf{RG+RR+Grouping \& Selection(GS)} &26.28 (1577) \\

\textbf{RG+RR+GS+Classification} &34.73 (2084) \\ \hline

\end{tabular}
}
\caption{Training time [minutes (seconds)] of IRIS ($50\%$) and baselines on VAST with zero-shot targets.}
\label{tab_train_time}
\end{table}

\section{Computation Time \label{app_train_time}}
Table \ref{tab_train_time} presents the training times of different stages of IRIS (50\%) compared to baseline methods. Since all baselines leverage external resources—such as Wikipedia or LLMs—for information retrieval, similar to our rationale generation stage, we treat these steps as preprocessing and exclude them from the reported training times for fairness. All models were run on our GPU setup, as detailed in Section \ref{sec_impl}). 
While IRIS generally requires more training time compared to other baselines (with comparable runtime to the Infuse method), it consistently delivers superior performance. This highlights a reasonable trade-off between computational cost and predictive effectiveness. Additionally, the training time of IRIS can be further optimized through parallel processing or more powerful GPU infrastructure—options we intentionally limited due to server availability constraints and our commitment to reducing carbon emissions.
\begin{figure}
\centering
\includegraphics[width=0.50\linewidth]{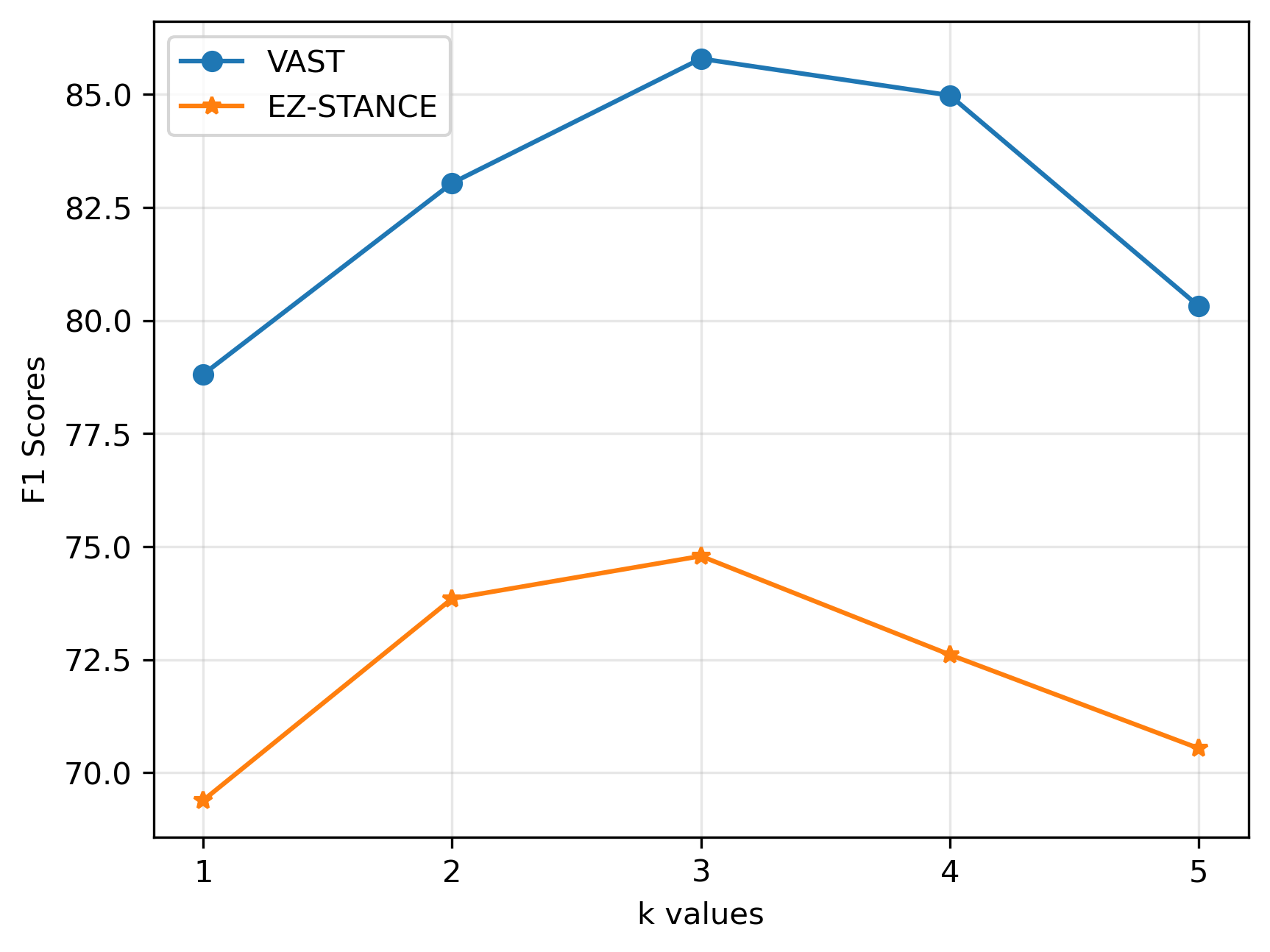}
 \caption{Different `k' values Performance for ZSSD}
  \label{fig_k1}
\end{figure}
\begin{figure}
\centering
\includegraphics[width=0.50\linewidth]{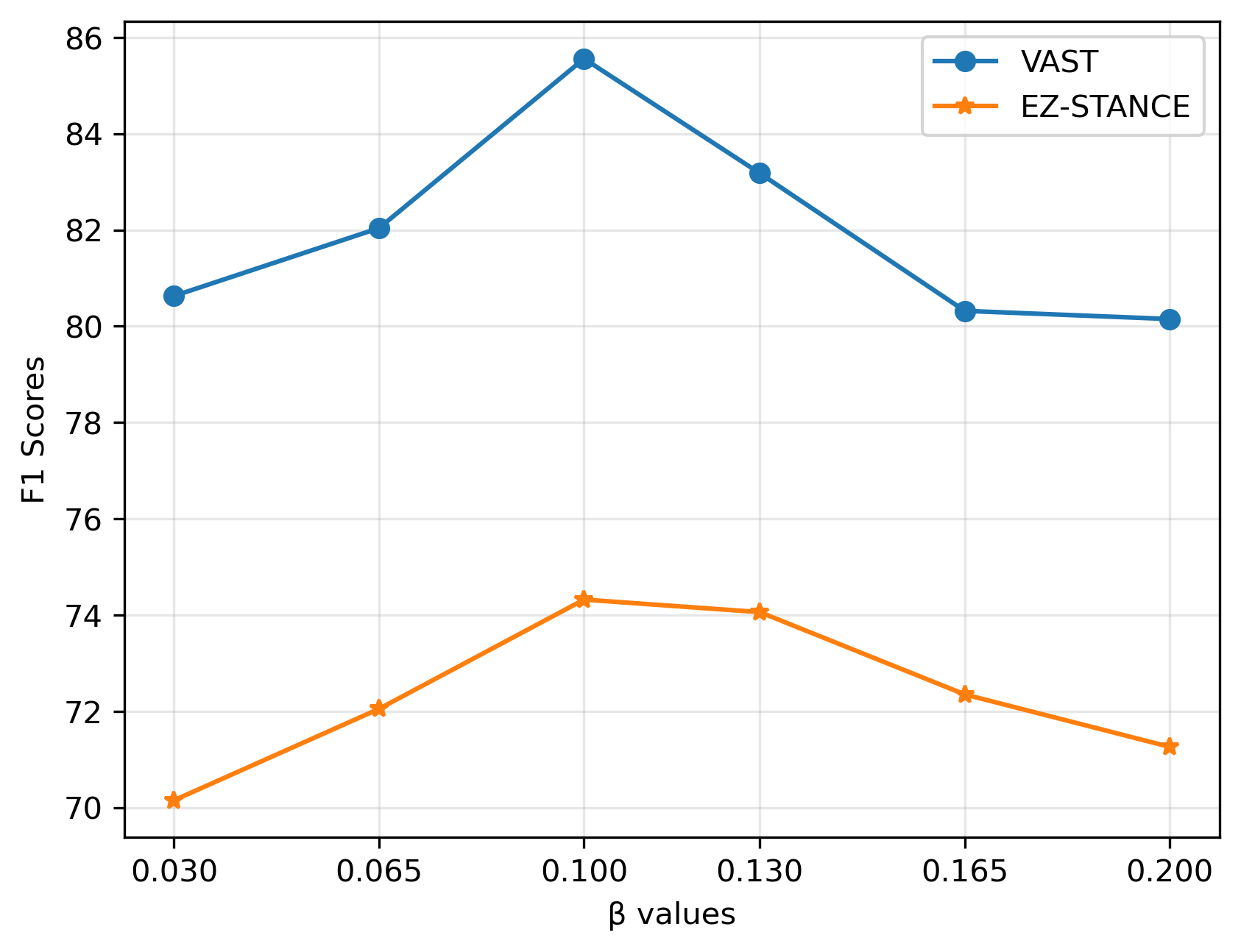}
 \caption{Different `$\beta$' values Performance for ZSSD}
  \label{fig_beta1}
\end{figure}

\begin{table}
\centering
\scalebox{0.70}{
\begin{tabular}{|l|l||l|}
\hline
\textbf{Model} & \textbf{VAST} &\textbf{EZ} \\ \hline
\textbf{{\begin{tabular}[c]{@{}l@{}}NV-Embed-v2\end{tabular}}} &85.56/0.33 &74.32/1.02 \\\hline 

\textbf{{\begin{tabular}[c]{@{}l@{}}bge-base-en-v1.5\end{tabular}}} &83.21/0.45 &71.66/0.29 \\ \hline

\textbf{{\begin{tabular}[c]{@{}l@{}}
stella\_en\_1.5B\_v5\end{tabular}}} &83.09/2.05 &72.31/2.41 \\ \hline

\textbf{{\begin{tabular}[c]{@{}l@{}}Sentence-BERT\end{tabular}}} &79.63/2.53 &68.28/2.39 \\ \hline

\textbf{{\begin{tabular}[c]{@{}l@{}}BART\end{tabular}}} &75.50/3.17 &65.45/3.03 \\ \hline

\textbf{{\begin{tabular}[c]{@{}l@{}}BERT\end{tabular}}} &76.36/3.09 &66.18/3.20 \\ \hline

\end{tabular}
}
\caption{Results (Average/Std. Dev) for different embeddings on VAST (zero-shot) and EZ (noun-phrase) for ZSSD.}
\label{tab_ablation_emb}
\end{table}

\section{Different Embeddings \label{app_embed}} We evaluate our approach with different sentence embeddings using HuggingFace MTEB Leaderboard\footnote{https://huggingface.co/spaces/mteb/leaderboard}, recent works \cite{lee2024nv} and other popular word embeddings. NV-Embed-v2 (https://huggingface.co/nvidia/NV-Embed-v2) leverages a decoder-only Mistral LLM with latent attention pooling and a high embedding dimension 4096. bge-base-en-v1.5 (https://huggingface.co/BAAI/bge-base-en-v1.5) is widely used for retrieval and ranking tasks, offering robust support for longer texts with a 768-dimensional embedding. "stella\_en\_1.5B\_v5" (https://huggingface.co/NovaSearch/stella\_en\_1.5B\_v5) is based on Alibaba-NLP’s gte-large-en-v1.5 and gte-Qwen2-1.5B-instruct huggingface models, providing embeddings with a default dimension of 1024. We also evaluate popular baseline embeddings such as Sentence-BERT \cite{reimers2019sentence}, BART \cite{lewis2019bart}, and BERT \cite{devlin2019bert}. It can be observed from Table \ref{tab_ablation_emb} that "NV-Embed-v2" embeddings excel at capturing underlying relationships more effectively than other embeddings. They demonstrate superior performance in identifying implicit relationships, subtle semantic nuances, and contextual cues, which are critical for tasks such as stance detection.

\begin{table*}

\centering

\scalebox{0.50}{
\begin{tabular}
{|l|p{0.60\linewidth}|p{0.15\linewidth}|p{0.30\linewidth}|p{0.70\linewidth}|}
\hline
S.No. &{\textbf{Text+Target}} &\textbf{Stance}  &\textbf{Implicit Rationale}   &\textbf{Explicit Rationale}\\ \hline

1. & \textbf{VAST:} If any of the dire prediction of a Trump presidency turning into a fascist dictatorship should start to occur... I do not find Justice Ginsburg's comments any less dignified..justices were to condemn Ms. Clinton's trustworthiness over her lies... The dignity of the court.. is lessened when..political maelstrom.\newline \textbf{Target:} Justice Ginsburg &\textbf{True:} against\newline \textbf{Predicted:} against 
 & ``less dignified than if one of the more conservative justices were to condemn Ms. Clinton's trustworthiness"; ``The dignity of the court, as well as the appearance of impartiality, is lessened when the justices lower themselves into the political maelstrom"; ``I do not find Justice Ginsburg's comments any less dignified"
 & The post exhibits low empathy, as it criticizes Justice Ginsburg's comments without considering her perspective or feelings. The language is also relatively concrete, focusing on specific events and actions rather than abstract ideas. There is a moderate level of agency language, as the author argues for justices to speak out and take action. However, there is no apparent allure or absolutist language in the post.
\\ \\\hline
2. & \textbf{EZ:} My girl is a dual Saudi American. In women s rights activists get detained tortured see: LoujainHathloul. We escaped saudi made it to the USA in 2019. My daughters teacher texted me today saying she is leading girls rights protests at school I AM SO PROUD!!!!\newline \textbf{Target:} women's rights activists  &\textbf{True:} favor\newline \textbf{Predicted:} favor	
 & ``We escaped saudi made it to the USA in 2019"; ``she is leading girls rights protests"; ``I AM SO PROUD"
 & The post exhibits high levels of Emotion (specifically, pride and concern), Agency\_language (the author is proud of their daughter taking action), and Concrete language (specific events and actions are described). There is also a sense of Approach language, as the author is motivated by positive emotions towards their daughter's actions. The tone is enthusiastic and celebratory.
\\\\ \hline

3. & \textbf{RFD:}When I got admitted to Yale...African-Americans rarely end up in elite institutions...Racism is not a zero-sum game, but opportunities for limited resources are...Given how much worse off African-Americans are than whites...President Obama’s “yes, we can” fervor seems to be contagious... affirmative action, U.S. Supreme Court Justice Antonin Scalia wrote “there are no debtor or creditor races.” But he was wrong. African-Americans still are owed big time. \newline \textbf{Traget:} Anti-White Bias Is a Problem.
 & \textbf{True:} against\newline \textbf{Predicted:} against
 & ``But, lest anyone worry, white folks, comparatively speaking, are doing just fine"; ``Blacks are twice as likely to be unemployed"; ``Six African-Americans head Fortune 500 companies. Of those businesses, 480 are run by whites"; ``We have one black president, and almost one million black people in prison"; ``African-Americans still are owed big time."; ``the most surprising thing is that many blacks believe there has been so much racial progress"
 & The post exhibits low empathy, as it challenges the target statement and presents a opposing view. \newline The language used has a high allure, aiming to persuade readers of the author's perspective.\newline Absolutist language is present, with statements like \"Racism is not a zero-sum game\".\newline The text promotes action, encouraging readers to acknowledge and address racism.\newline Abstract ideas are discussed, such as the concept of racism and its impact on society.\newline Concrete examples, like statistics on unemployment and prison rates, support the author's claims.\newline The language used is more agentic, with the author presenting their personal perspective and experiences.\newline Overall, the post takes an approach of confronting and challenging the target statement.
\\ \\ \hline

4. & \textbf{EZ:} On June 13, 1611, Smolensk fell to the Poles after the Russians defended it for 20 months. This occurred during the 1598-1613 Time of Troubles when Poland took advantage of Russia's weakness by engaging in regime change and military campaigns such as the occupation of Moscow.\newline\textbf{Target:} Smolensk
 & \textbf{True:} neutral\newline \textbf{Predicted:} \textcolor{red}{favor}
 & ``On June 13, 1611, Smolensk fell to the Poles"; ``Russians defended it for 20 months"; ``Poland took advantage of Russia's weakness
 & The post demonstrates Concrete language through historical specifics like "Smolensk fell," "20 months," and "occupation of Moscow." There's limited Empathy or Communion\_language, as the tone is factual and detached. Absolutist and Allure elements are absent, however, a little focus is on invoking motivation or ideology.
\\ \\ \hline

5.  & \textbf{VAST:} People used to have to have children in their teens and 20s because people didn't live very long...Maybe it is just natural part of human evolution as people have become more educated and science..\newline \textbf{Target:} age childbirth

 & \textbf{True:} neutral\newline \textbf{Predicted:} neutral
 & \textcolor{red}{Missing Relevant Rationales:} \textbf{by IRIS:} ``people didn't live very long"; `now people live a lot longer so there is no hurry"; `science has solved many of the problems of dieing off"
 \textbf{by Human:} ``people live a lot longer so there is no hurry";  ``many died in childhood, modern medicine has put an end to the need for that"; ``science has solved many of the problems of dieing off before life could be fully lived"
 & Based on the linguistic characteristics of the post, the post has low Empathy and Absolutist language, moderate Abstract and Concrete language, and high Agency\_language and Approach language. The tone is informative and neutral, with no emotional appeal or attempt to persuade. The language suggests a focus on individual perspective and personal goals rather than social relationships or collective well-being. 
\\ \hline
\end{tabular} 
}
\caption{Case studies of IRIS stance prediction including stance labels and rationales extraction.}
\label{tab_qual_exam}
\end{table*}
\section{Qualitative Analysis \label{app_qual}}
To understand the benefits of our prompts and IRIS approach, we looked into step-by-step processing for one of the sample predictions from the VAST dataset by LLMs only and our model: 
\textit{Target:} ``Justice Ginsburg"; \textit{Text:} ``If any of the dire prediction of a Trump presidency turning into a fascist dictatorship should start to occur... I do not find Justice Ginsburg's comments any less dignified..justices were to condemn Ms. Clinton's trustworthiness over her lies... The dignity of the court.. is lessened when..political maelstrom." \\ 
\textit{True Stance:} ``Against";\\
\textit{LLM stance prediction} (Llama 3.1 zero/few/fine-tuned): ``Neutral"
\par However, when asked Llama using prompt \ref{fig_prompt_im} to extract implicit rationales and prompt \ref{fig_prompt_ex} for linguistic measures, we obtain the below implicit and explicit rationales:\\
\textit{Implicit rationales ($IR_i$):} [$IR_1$: ``I do not find Justice Ginsburg's comments any less dignified", $IR_2$: ``less dignified than if one of the more conservative justices were to condemn Ms. Clinton's trustworthiness", $IR_3$: ``The dignity of the court, as well as the appearance of impartiality, is lessened when the justices lower themselves into the political maelstrom.", $IR_4$: ``than if one of the more conservative justices were to condemn", $IR_5$: ``I do not find Justice Ginsburg's comments"]\\
%\textbf{Corresponding Predictions for Implicit Rationales:} [["0.85", "0.05", "0.10"], ["0.05", "0.75", "0.20"],["0.10", "0.80", "0.10"],["0.3", "0.5", "0.2"],["0.4", "0.3", "0.3"]]\\
\textit{Explicit rationale ($ER$):} ``The post exhibits low empathy, as it criticizes Justice Ginsburg's comments without considering her perspective or feelings. The language is also relatively concrete, focusing on specific events and actions rather than abstract ideas. There is a moderate level of agency language, as the author argues for justices to speak out and take action. However, there is no apparent allure or absolutist language in the post"

\par  After passing through \textit{relevance ranking} stage:\\
\textit{Corresponding ranking scores for implicit rationales $\{S^{rf}_i, S^{ra}_i, S^{rn}_i\}$:} [[``0.85", ``0.05", ``0.10"], 
 [``0.05", ``0.75", ``0.20"], [``0.25", ``0.55", ``0.20"], [``0.3", ``0.2", ``0.5"], [``0.4", ``0.3", ``0.3"]]
\par \textit{Grouping and selection} stage result in the following set of relevant rationales:\\
\textit{Relevant rationales:}$IR_2$: ``less dignified than if one of the more conservative justices were to condemn Ms. Clinton's trustworthiness", $IR_3$: ``The dignity of the court, as well as the appearance of impartiality, is lessened when the justices lower themselves into the political maelstrom"; $IR_5$: ``I do not find Justice Ginsburg's comments any less dignified"\\
\textit{Stance prediction of the concatenation of each relevant implicit and explicit implicit:}($IR_1, ER$): Against; ($IR_2, ER$): Against; ($IR_3, ER$): Favor\\
\textit{Final Stance by IRIS (majority vote):} ``Against" 
\par This justifies that our IRIS was able to focus on the relevant rationales and the concatenation of explicit and implicit rationales followed by a majority vote leads to a correct prediction of the attitude towards the given target.

\end{document}